\documentclass[10pt,twocolumn,letterpaper]{article}

\usepackage[pagenumbers]{iccv} 

\usepackage{longtable}
\usepackage{comment}
\usepackage{booktabs}
\usepackage{graphicx}
\usepackage{multirow}
\usepackage{fixmath,mathtools,nicefrac}
\usepackage{url}            
\usepackage{booktabs}       
\usepackage{amsfonts}       
\usepackage{nicefrac}       
\usepackage{microtype}      
\usepackage{makecell}
\usepackage{adjustbox}
\usepackage{pifont}
\usepackage{bbding}
\usepackage{enumerate}
\usepackage{enumitem}
\usepackage{subcaption}
\usepackage{comment}
\usepackage{caption}
\usepackage{float}
\usepackage{colortbl}
\usepackage{wrapfig} 
\usepackage[misc]{ifsym} 
\usepackage{pifont}
\usepackage{arydshln}
\usepackage{tcolorbox}

\newcommand{\pipelinename}{MM-IFEngine\xspace}
\newcommand{\datasetname}{MM-IFInstruct-23k\xspace}
\newcommand{\dpodatasetname}{MM-IFDPO-23k\xspace}
\newcommand{\benchmarkname}{MM-IFEval\xspace}

\definecolor{lightblue}{HTML}{DAEFF9}
\definecolor{bisque}{rgb}{1.0, 0.89, 0.77}

\usepackage{xcolor}
\usepackage{graphicx}

\definecolor{ForestGreen}{rgb}{0, 0.69, 0.31}
\definecolor{NavyBlue}{rgb}{0, 0.44, 0.75}

\newcommand{\hgreen}[1]{\textcolor{ForestGreen}{\textbf{#1}}} 

\definecolor{iccvblue}{rgb}{0.21,0.49,0.74}
\usepackage[pagebackref,breaklinks,colorlinks,allcolors=iccvblue]{hyperref}

\def\paperID{4867} 
\def\confName{ICCV}
\def\confYear{2025}

\title{MM-IFEngine: Towards Multimodal Instruction Following
}

\author{\bf Shengyuan Ding$^{1,2*}$, Shenxi Wu$^{1,2*}$, Xiangyu Zhao$^{2,3}$, Yuhang Zang$^{2}\textsuperscript{\Letter}$, \\ \bf Haodong Duan$^{2}$,
Xiaoyi Dong$^{2}$, Pan Zhang$^{2}$, Yuhang Cao$^{2}$, Dahua Lin$^{2,4,5}$, Jiaqi Wang$^{2,6}\textsuperscript{\Letter}$
\\
$^1$Fudan University \quad
$^2$Shanghai AI Laboratory \quad
$^3$Shanghai Jiaotong University \\
$^4$The Chinese University of Hong Kong \quad
$^5$CPII under InnoHK \quad
$^6$Shanghai Innovation Institute
}

\begin{document}

\twocolumn[{%
\renewcommand\twocolumn[1][]{#1}%
\maketitle

\begingroup
\renewcommand\thefootnote{}\footnote{* Equal contribution. \Letter\ Corresponding authors.}
\addtocounter{footnote}{-1}
\endgroup

\vspace{-10mm}
\begin{center}
    \centering
    \captionsetup{type=figure}
    \includegraphics[width=\linewidth]{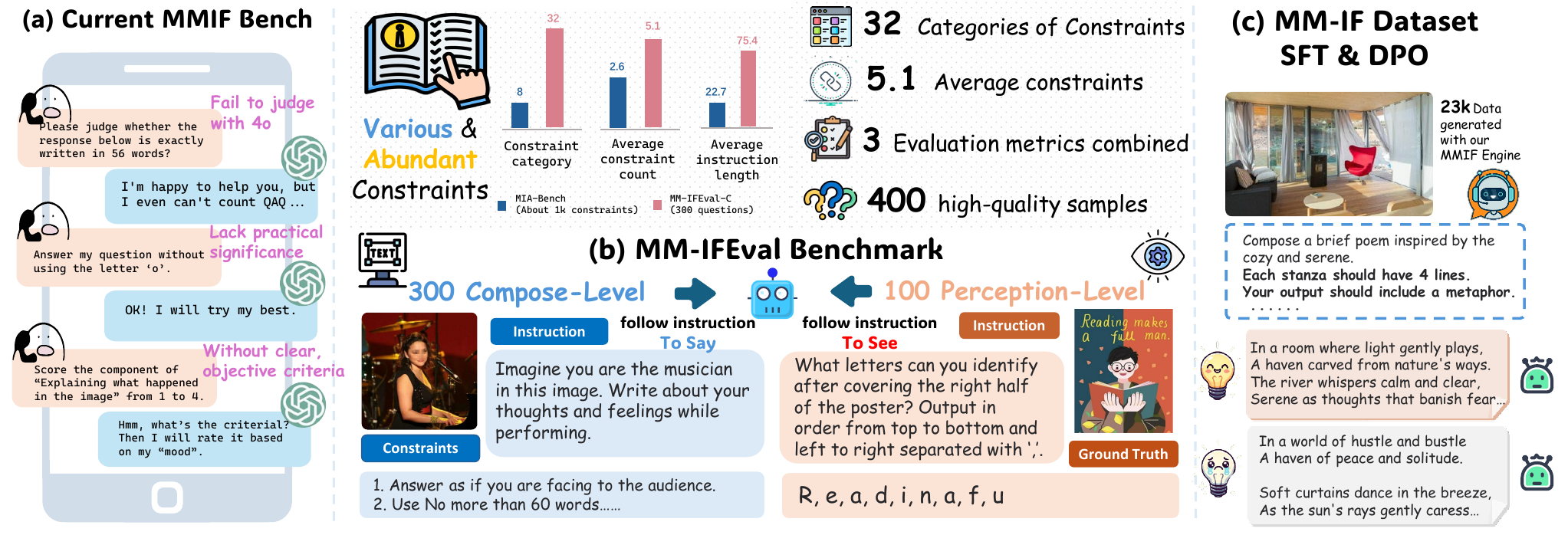}
    \caption{
    \textbf{(a)} Limitations of existing Multimodal Instruction Following (IF) benchmarks.
    \textbf{(b)} Overview of the MM-IFEval benchmark, which significantly surpasses existing benchmarks in terms of constraint diversity, quantity, and instruction complexity. Our benchmark consists of Compose-Level (C-Level) problems that impose constraints on model outputs (e.g., format requirements, keyword limits) and Perception-Level (P-Level) problems that require reasoning about specific visual elements in images.
    \textbf{(c)} Our \pipelinename generates a large-scale, diverse training dataset suitable for both Supervised Fine-Tuning (SFT) and Direct Preference Optimization (DPO).
    }
    \label{fig:teaser}
\end{center}%
}

]


\begingroup
\renewcommand\thefootnote{}\footnote{* Equal contribution. \Letter\ Corresponding authors.}
\addtocounter{footnote}{-1}
\endgroup

\begin{abstract}
The Instruction Following (IF) ability measures how well Multi-modal Large Language Models (MLLMs) understand exactly what users are telling them and whether they are doing it right.
Existing multimodal instruction following training data is scarce, the benchmarks are simple with atomic instructions, and the evaluation strategies are imprecise for tasks demanding exact output constraints.
To address this, we present MM-IFEngine, an effective pipeline to generate high-quality image-instruction pairs.
Our MM-IFEngine pipeline yields large-scale, diverse, and high-quality training data MM-IFInstruct-23k, which is suitable for Supervised Fine-Tuning (SFT) and extended as MM-IFDPO-23k for Direct Preference Optimization (DPO).
We further introduce MM-IFEval, a challenging and diverse multi-modal instruction-following benchmark that includes (1) both compose-level constraints for output responses and perception-level constraints tied to the input images, and (2) a comprehensive evaluation pipeline incorporating both rule-based assessment and judge model.
We conduct SFT and DPO experiments and demonstrate that fine-tuning MLLMs on MM-IFInstruct-23k and MM-IFDPO-23k achieves notable gains on various IF benchmarks, such as MM-IFEval (+10.2$\%$), MIA (+7.6$\%$), and IFEval (+12.3$\%$).
We have fully open-sourced the datasets (both SFT and DPO), evaluation code and training scripts at \url{https://github.com/SYuan03/MM-IFEngine}.
\end{abstract}    
\section{Introduction}
Instruction Following (IF) is a fundamental ability in Large Language Models (LLMs) \cite{lou2023comprehensive,jiang2023followbench,qin2024infobench,zhang2024cfbench,zhou2023instruction} and Multimodal Large Language Models (MLLMs) \cite{bitton2023visit,qian2024mia}, which involves accurately interpreting and executing user-provided instructions.
This ability is crucial for deploying models in real-world applications where users expect precise and context-aware responses, such as code generation \cite{xu2023wizardlm}, visual question answering \cite{li2024textbind}, robots \cite{shi2025hi}, and creative content creation \cite{zhou2023controlled}.
For instance, in a VQA scenario, when a user asks an MLLM \texttt{what is the object and how do I use it, return the object name and the usage instructions in a JSON format}, accurate IF ensures the model provides a response like \{\texttt{object}': `\texttt{hammer}', `\texttt{usage}': `\texttt{use it to drive nails}'\} instead of the plain text. 

Achieving precise IF in multimodal, diverse, and open-ended environments presents significant challenges for both \textit{model training} and \textit{benchmark evaluation}.
One significant limitation is the scarcity of high-quality IF training data to train open-source MLLMs.
In addition, current multimodal IF benchmarks \cite{bitton2023visit,qian2024mia} merely have simple, atomic instructions, and the constraints are weakly correlated with visual content (see \cref{fig:teaser} \textbf{(a)}).
Consequently, existing benchmarks lack the diversity required for real-world applications, leading to saturated results where nearly all models achieve over 80\%.
Furthermore, the evaluation method in existing benchmarks often relies on LLM-as-a-judge \cite{zheng2023judging}, which is imprecise for instructions demanding exact output constraints, such as word counts.
Therefore, the combination of \textit{limited training data}, \textit{simple benchmarks}, and \textit{imprecise evaluation strategy} strongly restricts the progress of current MLLMs in IF.

To address the lack of high-quality IF training data and challenging benchmarks, we propose \textbf{\pipelinename}, an effective pipeline for generating high-quality image-instruction pairs.
\pipelinename collects diverse image sources, including natural scenes, UI interfaces, diagrams, charts, and mathematical problems.
We then employ a structured approach using a predefined set of 16 task descriptions and 32 constraints to guide the LLM in crafting tailored instructions for each image.
Using \pipelinename, we generated a comprehensive dataset of image-instruction pairs, collected responses from open-source MLLMs, and applied rigorous post-processing to retain only high-quality instruction-answer pairs, thus constructing \textbf{\datasetname} for Supervised Fine-Tuning (SFT).
We also generate negative responses by selectively removing constraints from the original data, constructing the preference dataset \textbf{\dpodatasetname} for preference optimization algorithms such as Direct Preference Optimization (DPO) \cite{rafailov2023direct}.

To facilitate the evaluation of multimodal IF, we present \textbf{\benchmarkname}, a benchmark comprising 400 challenging problems with diverse compose-level and perception-level instructions.
\benchmarkname is derived from the images and instructions generated by \pipelinename with human-labeled annotations.
As presented in \cref{fig:teaser} \textbf{(b)}, our \benchmarkname has the following three distinctive features: (1) \textbf{Diverse Instruction Types}: \benchmarkname has 32 distinct constraints, ensuring a wide range of instruction complexities and surpassing the scope of prior benchmarks. (2) \textbf{Hybrid Evaluation}: we use a hybrid strategy including both rule-based verification and judge model.
For subjective instructions (e.g., mimicking tone), we design a \textit{comparative} judgment for precise evaluation. Specifically, a control output is generated without the constraint, and the LLM judge compares both outputs for precise evaluation.
(3) \textbf{Challenging}: the leading proprietary model (GPT-4o at 64.6\%) and open-source model (Qwen2-VL-72B at 50.8\%) demonstrating substantial room for improvement on our benchmark, highlights a significant opportunity for improvement in multimodal instruction following.

We further demonstrate that fine-tuning MLLMs on either \datasetname or \dpodatasetname consistently boosts the performance of MLLMs on instruction following benchmarks, without compromising their original capabilities on other Visual Question Answering (VQA) benchmarks.
Specifically, fine-tuning Qwen2-VL-7B on \dpodatasetname with the DPO results in performance gains of 10.2$\%$, 7.6$\%$, and 12.3$\%$ on \datasetname, MIA-Bench \cite{qian2024mia}, and IFEval \cite{zhou2023instruction}, respectively.

Our contributions include: \textbf{(1)} a \pipelinename pipeline for generating multimodal constraint-rich image-instruction pairs;
\textbf{(2)} a large-scale training dataset \datasetname and preference optimization dataset \dpodatasetname derived from \pipelinename;
\textbf{(3)} a challenging multimodal instruction following benchmark \benchmarkname with diverse constraints and comprehensive evaluation approaches;
and \textbf{(4)} empirical evidence showing significant performance gains on both our \benchmarkname and existing benchmarks when training MLLMs on \datasetname via SFT and \dpodatasetname via DPO.
\section{Related Work}

\begin{figure*}[t]
\centering
\includegraphics[width=\linewidth]{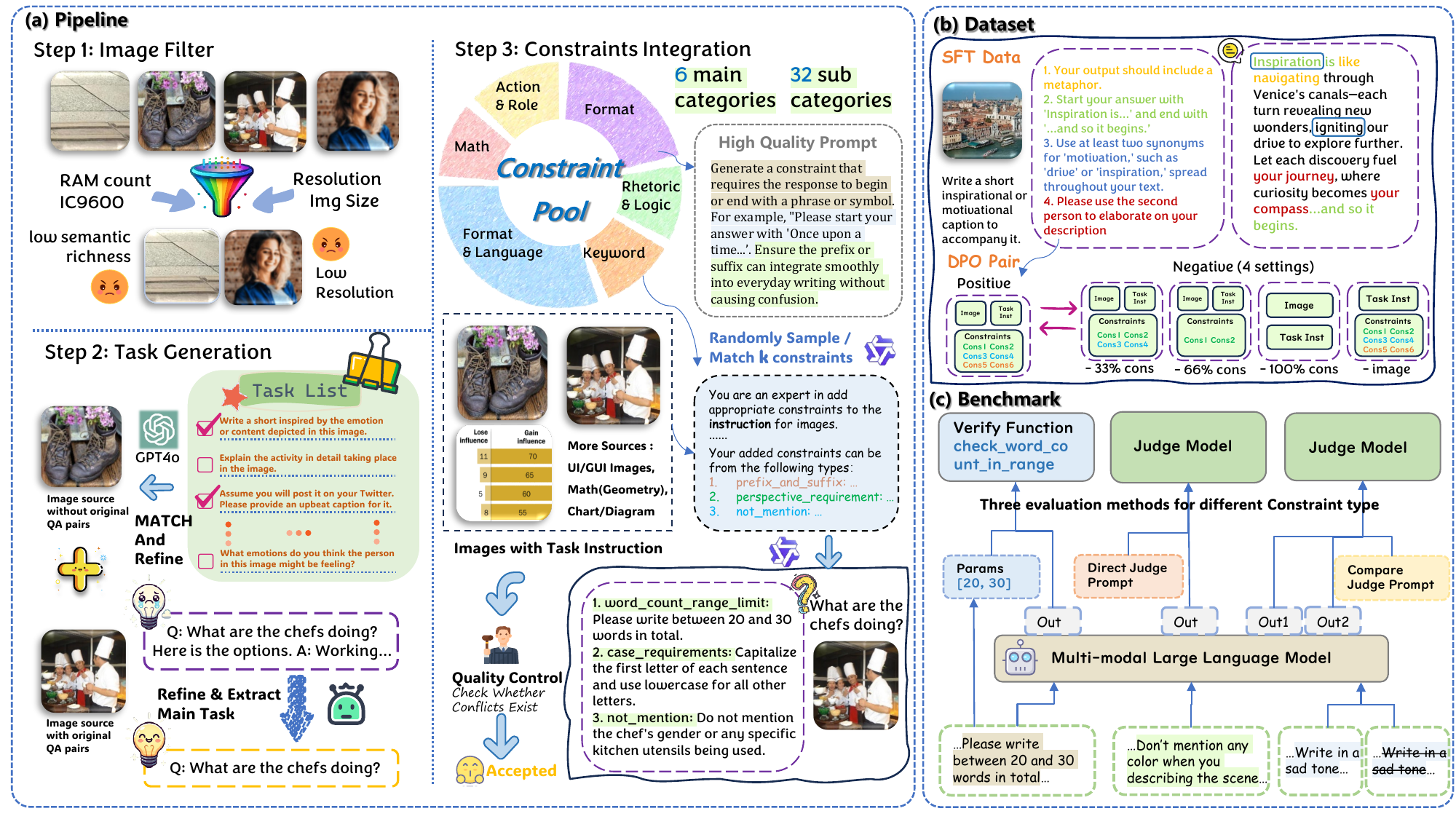}
\caption{\textbf{Overall pipeline of MM-IFEngine}. Part \textbf{(a)} demonstrates the three-stage workflow of our engine: (1) Image filter; (2) Task generation using GPT-4o for images without QA pairs and instruct refinement for existing annotations; and (3) Constraints integration incorporating 6 main categories and 32 subcategories, ensuring compatibility between constraints and tasks. MM-IFEngine is employed to generate SFT and DPO training datasets and MM-IFEval benchmark, as shown in part \textbf{(b)} and \textbf{(c)}. MM-IFEval implements three evaluation metrics combining rule-based verification functions and a judge model to ensure accurate assessment.
}
\label{fig:pipeline}
\end{figure*}

\textbf{Instruction Following in LLMs}.
Various benchmarks and training approaches have been proposed to make Large Language Models (LLMs) better align with human instructions.
While existing Instruction Following (IF) benchmarks like \cite{zhou2023instruction,jiang2023followbench,zhang2024cfbench,qin2024infobench} all aim to evaluate instruction following, they differ significantly in their \textit{dataset construction pipelines}, driven by their unique constraint taxonomies.
CFBench \cite{zhang2024cfbench}, for instance, constructs its dataset using a combination of taxonomic and statistical methodologies to establish comprehensive constraints.
This divergence extends to their \textit{evaluation strategies}.
For example, InFoBench \cite{qin2024infobench} adopts a strategy of decomposing complex instructions into simpler assessment standards.
Beyond benchmarks, various training approaches aim to enhance LLMs' instruction-following capabilities \cite{xu2023wizardlm,luo2023wizardcoder}, including in-context learning \cite{zhou2023controlled} and preference optimization \cite{zhang2024iopoempoweringllmscomplex}.
However, he aforementioned research is limited to the text modality, whereas our work focuses on multi-modal instruction following with vision inputs.
 
\noindent\textbf{Instruction Following Benchmarks in MLLMs}.
Numerous benchmarks \cite{li2024survey} have been proposed to evaluate diverse capabilities of Multi-modal Large Language Models (MLLMs), including general knowledge \cite{yue2024mmmu,liu2024mmbench,chen2024we,mmtbench}, document understanding \cite{kembhavi2016diagram,liu2024ocrbench,ma2024mmlongbench}, perception \cite{zang2025contextual,wei2025videorope}, multi-image comprehension \cite{liu2024mmdu,song2024milebenchbenchmarkingmllmslong,wang2024muirbenchcomprehensivebenchmarkrobust}, and instruction following (IF) \cite{bitton2023visit,qian2024mia}.
MIA-Bench~\cite{qian2024mia} and VisIT-Bench \cite{bitton2023visit} are representative IF benchmarks that employ GPT-4 \cite{openai2023gpt4} for question generation and evaluation.
In contrast to existing IF benchmarks, our \benchmarkname introduces significant improvements in diversity (32 constraint categories covering compositional and perceptual aspects), difficulty (averaging 5.1 constraints per question), and evaluation precision (using both judge models and rule-based verification).

\noindent\textbf{Instruction Tuning Data for MLLMs}.
Recent advancements in multi-modal instruction tuning data aim to improve cross-modal alignment and increase the variety of tasks handled by MLLMs \cite{liu2023visual,chen2023sharegpt4vimprovinglargemultimodal,dai2023instructblipgeneralpurposevisionlanguagemodels,xu2023multiinstructimprovingmultimodalzeroshot,xu2024visionflanscalinghumanlabeledtasks,liu2024mmdu, zang2025internlm}.
For example, some previous works \cite{chen2023sharegpt4vimprovinglargemultimodal,chen2024allava,Liu_2024} build synthetic instruction tuning data generated using GPT-4V~\cite{openai2023gpt4v}, enabling open-source MLLMs to achieve performance comparable to proprietary models across multiple benchmarks.
However, existing instruction tuning data are mainly designed for general knowledge or visual perception, and data for improving the IF abilities is scarce.
The scarcity of training data for enhancing IF abilities motivated the development of our \pipelinename pipeline.

\section{\pipelinename}\label{sec:pipeline}

We employ the \pipelinename pipeline to generate image-instruction pairs, which are the foundation for creating instruction tuning data and our benchmark.
As shown in \cref{fig:pipeline} \textbf{(a)}, the pipeline is composed of three main stages: (1) image filtering, where we systematically select a diverse set of images from multiple sources to ensure broad coverage of visual content; (2) task generation, in which we either synthesize novel tasks tailored to the selected images or refine existing instruction templates to better align with the image content; and (3) constraint integration, where high-quality, constraint-aware instructions are generated for images that initially lack associated annotated guidance, thereby enhancing the richness and precision of the dataset.

\subsection{Image Filter}
Our image filtering strategy selects only high-quality images by removing those with low resolution or limited semantic richness.
For unannotated pure image datasets (e.g., CC3M \cite{sharma2018conceptual}), we prioritize natural scene images. Rich semantic content in these images enables the creation of more comprehensive and insightful QA pairs, which is crucial for designing diverse and complex instruction following tasks.
We use the IC9600 and RAM metric proposed in the previous method \cite{zhao2025omnialign} to select the images that have rich semantic content.

Furthermore, we analyze existing annotated datasets, such as ALLaVA \cite{chen2024allava}.
Our analysis reveals that some images suffer from low resolution, making them inadequate for the instruction-following task.
Given our intention to design more intricate and varied instruction following tasks based on this data, we filter out data items containing low-quality images.

\subsection{Task Generation}
\noindent \textbf{Image Source without Original QA Pairs.} For image datasets lacking original annotated task instructions (e.g., CC3M \cite{sharma2018conceptual}), we first design appropriate task instructions for the data items.
We first develop a series of task instructions tailored to the data items. 
These instructions are crafted to elicit long-form responses that can be subsequently modified or refined using various constraints, for instance, \textit{Provide a detailed analysis of the image, including the setting, characters, and notable objects.} 
The final task pool $\mathcal{P}_T$ comprises a total of 16 distinct tasks, with further details available in Appendix \ref{app:task_pool}.


Given the task pool \(\mathcal{P}_T\), we randomly select \(k\) tasks as examples of task types for each image \(I\). We then prompt a powerful language model \(\mathcal{M}\) (e.g., GPT-4o) to generate an appropriate task list \(T_l\) that aligns with the image content. The process is formulated as:

\begin{equation}
    \{T^*_l\} = \mathcal{M}(I, T_e)
\end{equation} where \(T_e=\{T_1,T_2,\dots,T_k\}\) and each \(T_i\in\mathcal{P}_T\). The model \(\mathcal{M}\) is tasked with either choosing relevant tasks from \(T_e\) or supplementing reasonable tasks to construct the appropriate task list \(T^*_l\), ensuring that all tasks in \(T^*_l\) are in line with the image content. After generating the \(T^*_l\), a sampling step is incorporated to guarantee task diversity. For each image,  tasks are sampled. This sampling process is crucial as it enriches the variety of tasks associated with each image.

\noindent \textbf{Image Source with QA Pairs.} In the case of image datasets that have QA pairs (e.g., ALLaVA \cite{chen2024allava}), we adopt certain strategies for processing the original question annotations.
We choose ALLaVA as the primary dataset for this type of image source due to its rich and diverse image content, which is accompanied by a variety of task types. First, we conduct an analysis of the original question annotations. We find that some of the questions are accompanied by some few-shot examples. Additionally, some questions in ALLaVA have options in their original annotations, which are not suitable for our instruction-following task. Since we need to incorporate certain constraints into the original instructions in the subsequent steps, we use regular expressions and length limits to filter the questions in ALLaVA. Specifically, we select those questions that do not have few-shot examples associated with them. Mathematically, if we let \(Q\) be the set of all questions in ALLaVA, \(Q_{fs}\) be the subset of questions with few-shot examples, and \(Q_{op}\) be the subset of questions with options. We aim to find the subset \(Q_{s}\) of questions that satisfy the conditions:
\begin{equation}
    Q_{s}=\{q\in Q|q\notin Q_{fs}\land q\notin Q_{op}\}
\end{equation} where the filtering based on the absence of few-shot examples and options is achieved using regular expressions and length limits. Then, we get the expected \(T^*\) in our filter \(Q_{s}\) set for the images.


\subsection{Constraints Integration}
\noindent \textbf{Constraints Pool ($\mathcal{P}_C$)} \label{sec:pipeline_Taxonomy_of_Constraints}
We use \textit{instruction} to refer to the entire textual input, which in our paper can generally be viewed as a composition of \textit{a task instruction} and \textit{multiple constraints instruction}.
Tasks and constraints are rich and diverse, with a certain complexity in our work.
All the constraints in our work can be further classified into six major categories, each with its own unique characteristics and applications: Text Length Requirements, Mathematical Requirements, Language \& Formatting Requirements, Rhetoric \& Logic Requirements, Action Requirements, and Keyword Requirements.
Please refer to the Appendix \cref{fig:Taxonomy_of_Constraints} for more details of all the constraints.

Given the constraints pool $\mathcal{P}_C$ and task instructions, a straightforward approach for composing full instruction is to first set several constraints for each constraint type and then randomly select one constraint from some of the types to compose the constraint list, and finally concatenate the constraint list with the task instruction to form the full instruction. 
But this direct method has two problems: (1) The constraints are not diverse enough, which may not be able to fully evaluate the ability of the model. (2) The contradiction between the constraints and also between the constraints and the task instruction may exist.
For the first problem, an LLM is employed to generate concrete content of constraint instruction for the specific constraint type in our method. In order to avoid the generated content being too divergent or hard to control its difficulty, we carefully design some cases or requirements of details that needed to be paid attention to when generating the content for each constraint type (Appendix~\ref{app:constraints_category}). For the second problem, we also use a powerful LLM to help keep the correlation of constraints with its instruction and filter out those that cause total contradiction. Finally, we prompt an LLM to check whether the constraints and the task instruction are compatible and filter out those failing to pass the check. Our method not only ensures the compatibility of constraints and instructions but also enriches the diversity of constraints.

In our actual practice process, we find that although we prompt the LLM to select appropriate constraints that should be compatible with the task instruction and other constraints, the generated constraints still have some contradiction with the task instruction, especially on those existing datasets with various kinds of annotations. The reason is that these datasets are designed for overall question-answering tasks, and the question(or named task instruction) tends to be contradictory with the constraints, which are mostly compatible with those tasks of creating or answering in non-short form.
So, we decouple the selection and generation steps for this type of data source. Specifically, we first select the constraints from the constraints pool $\mathcal{P}_C$ and then provide the selected mostly compatible constraints to the LLM to select secondly and generate final constraints. But for image datasets without original QA pairs, in other words, for which we generate task instructions for them using $\mathcal{P}_T$, we directly sample k constraint types for the LLM to generate concrete content because they are mostly compatible with the pre-designed task instruction. The uniform process is formulated as:  
\begin{equation}
    C^*_l = \mathcal{L}(C_s, T^*), C^*_f = \mathcal{V}(C^*_l, T^*)
\end{equation}
where $\mathcal{T}^*$ is the task applicable to the image. The model \(\mathcal{L}\) is tasked with both choosing appropriate constraint types from \(C_s\) again and generating concrete constraints for some of them, whose output is a list of concrete constraint descriptions. To ensure that the generated constraints remain compatible with the given task instruction \(T^*\), we employ a final validation step using another LLM process, denoted as \(\mathcal{V}\). This validation function checks whether each constraint in \(C^*_l\) aligns with \(T^*\) and filters out those that contradict or do not fit the task instruction. The resulting set of fully verified and compatible constraints is represented as \(C^*_f\).  

\begin{figure*}[t]
\centering
\begin{minipage}{0.48\textwidth}
\centering
\includegraphics[width=\linewidth]{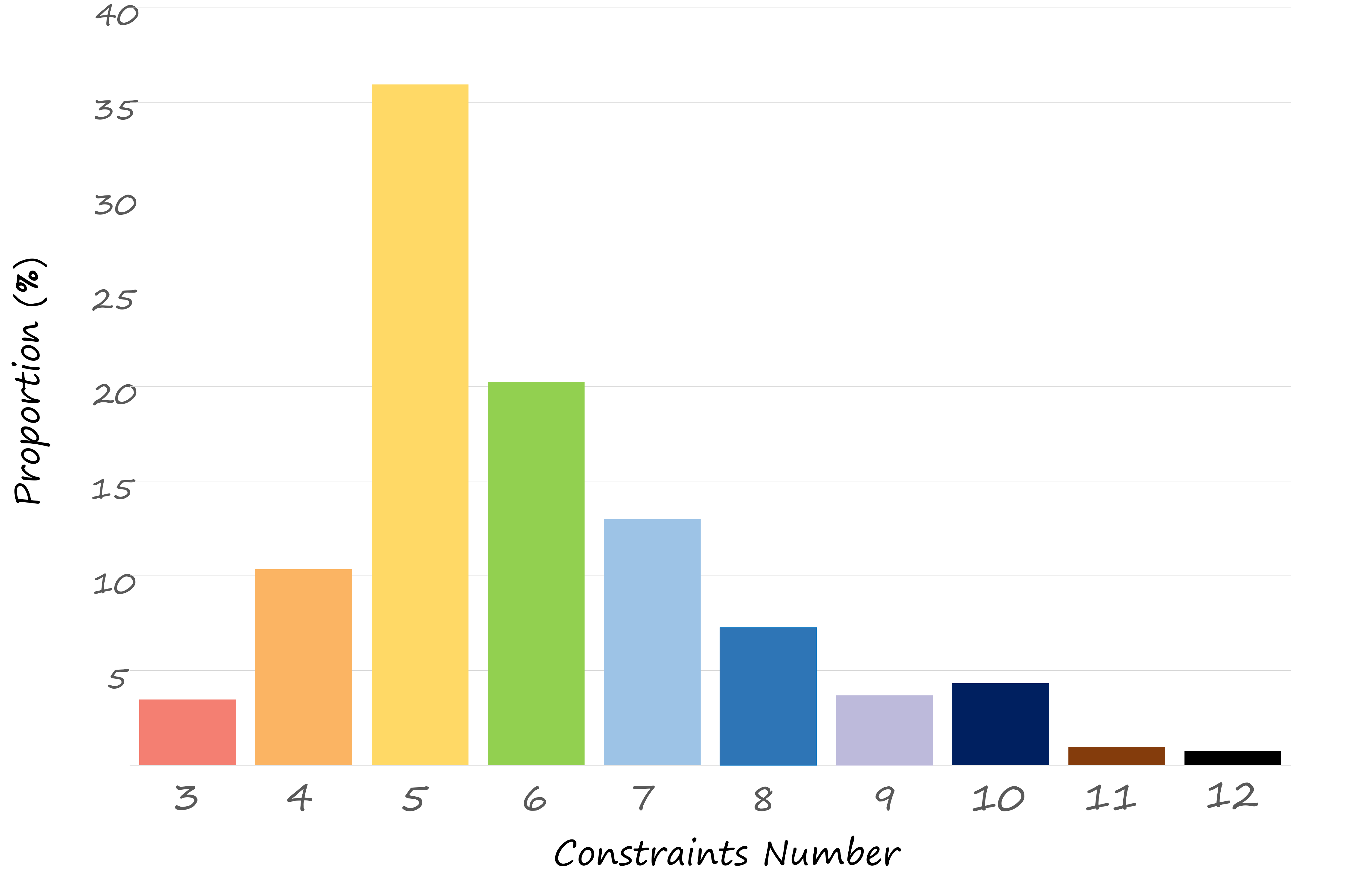}
\vspace{-6pt}
\caption{\textbf{Constraint Quantity Distribution in \datasetname}. Our \datasetname exhibits systematic variation in constraint complexity, with each sample containing 3-12 constraints per instruction.}
\vspace{-6pt}
\label{fig:statis}
\end{minipage}
\hspace{+4pt}
\begin{minipage}{0.48\textwidth}
\centering
\includegraphics[width=.7\linewidth]{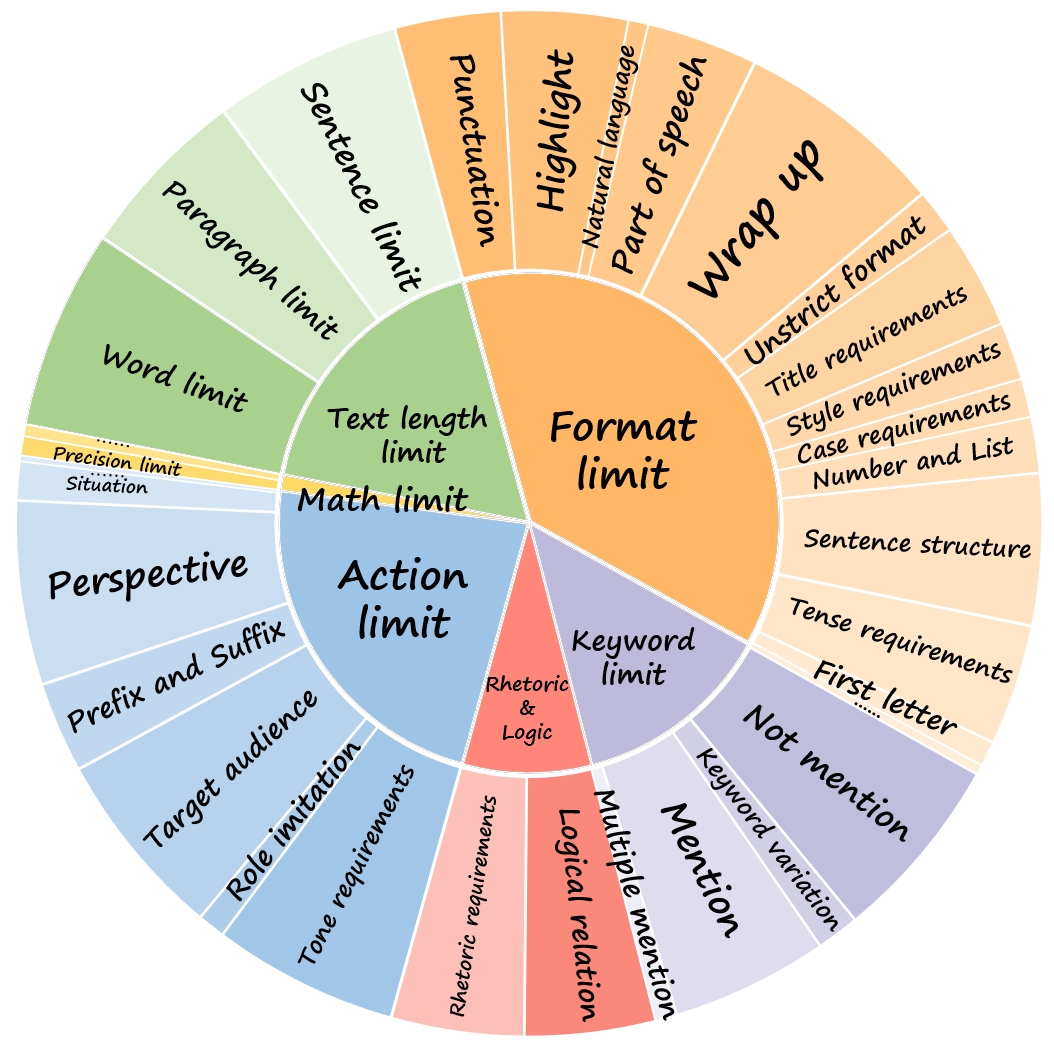}
\caption{\textbf{Constraint Category Distribution in Compose-Level Problems of MM-IFEval}. This part comprises six primary constraint categories with 32 subcategories, forming a multi-level taxonomy for instruction-following evaluation.
}
\label{fig:non_vision_subcategory}
\end{minipage}
\end{figure*}

\noindent\textbf{\datasetname Construction}.  
By applying the \pipelinename pipeline, we construct the \datasetname dataset, which contains 23k high-quality multi-modal instruction-following training data.
We first take an analysis of the performance of the current open-source MLLMs and proprietary MLLMs on several benchmarks \cite{qian2024mia, liu2024ocrbench}, and find that for instruction-following capability, the most powerful open-source MLLM like InternVL2.5-78B-MPO \cite{wang2024enhancing} is nearly equivalent to GPT-4o, and the performance on general VQA benchmarks are even higher then GPT-4o.
Thus, we use InternVL2.5-78B-MPO to generate responses for our \datasetname dataset.
Despite its capabilities, the InternVL2.5-78B-MPO model encounters difficulties in ensuring 100\% compliance with our constraints, a challenge attributed to the complexity, number, and comprehensiveness.
Consequently, we implement a post-processing stage to filter out responses that do not meet the specified criteria.
Acknowledging that achieving perfect constraint adherence might be challenging even for human annotators on this task, we set a practical accuracy threshold of 80\%.
Finally, our \datasetname comprises 23k data items, with 16k constructed from the training set of CC3M, 6k from ALLaVA, and 4k from the training set of MultiUI, Geo170k\cite{gao2023g} and ChartQA\cite{masry2022chartqa}.
We show the distribution of constraints number of \datasetname in \cref{fig:statis}.

\noindent\textbf{\dpodatasetname Construction}.
To comprehensively explore and make full use of our high-quality data, we also utilize \pipelinename to construct \dpodatasetname, a preference dataset comprising chosen and rejected samples suitable for Direct Preference Optimization (DPO) \cite{rafailov2023direct}.
Our high-quality data can be directly employed as the chosen samples.
Regarding rejected samples, we opt to utilize Qwen2-VL-7B-Instruct to answer the variant of the question for generating rejected pairs.
Specifically, we have four distinct settings for generating negative pairs, which mainly differ in the input to Qwen2-VL-7B-Instruct. These settings include  (1) With image, but randomly remove one-third of the number of constraints in the prompt; (2) With image, but randomly remove two-thirds of the number of constraints in the prompt; (3) With image, but randomly remove all the constraints in the prompt; and (4) Full prompt, but without the image;
We use these four types of input to feed into Qwen2-VL-7B-Instruct model, and collect the rejected responses to construct the \dpodatasetname.

\section{MM-IFEval}
Existing benchmarks for multi-modal instruction following are scarce. The majority focus on simple and atomic instructions, resulting in performance saturation across models. To address this limitation, we introduce \textbf{\benchmarkname}, a human-annotated, comprehensive, and challenging benchmark designed for evaluating multi-modal IF.

\subsection{\benchmarkname Construction}
To construct the \benchmarkname, we first use our \pipelinename to generate the question-answer (QA) pairs for images.
The generated instructions may inherently contain potential conflicts.
Consequently, human annotation remains critical for constructing this benchmark, as human annotators possess the cognitive capacity for comprehensive assessment of these complex situations.
After the human annotation, we further use an extra post-processing step that prompts the LLMs to double-check and mitigate the occurrence of constraint conflicts as much as possible.
Finally, we construct the \benchmarkname bench of 400 questions, 300 of which are \textit{compose-level} open-ended questions and 100 \textit{perception-level} questions with ground truth.

\noindent \textbf{Diverse Constraints.} With 32 distinct constraint categories and an average of 5.1 constraints per question, \benchmarkname presents a more challenging evaluation task compared to earlier benchmarks (e.g., \cite{qian2024mia}, which has 8 categories and 2.6 average constraints per question). Furthermore, our benchmark incorporates essential constraints such as ``Output in JSON format", which is prevalent and practical in real-world scenarios, a feature not found in previous multi-modal instruction following benchmarks.

\noindent \textbf{Compose-level and Perception-level Questions.} \textit{Compose-level} questions involve textual constraints, while \textit{perception-level} questions require greater visual perception ability to solve.
The perception-level questions incorporate a variety of image sources, such as natural scenes, user interfaces, diagrams, table charts, and mathematical expressions, which we believe are representative of real-world applications.
Please refer to the Appendix for examples of compose-level and perception-level questions.

\begin{table*}[ht]
    \centering
    \small
    \caption{\textbf{Main results on Instruction Following benchmarks}, including our proposed \benchmarkname, MIA-Bench \cite{qian2024mia}, and IFEval \cite{zhou2023instruction}. The symbol \textsuperscript{M} refers to multimodal benchmarks, and \textsuperscript{T} denotes text-only benchmarks. We report both compose-level (``C'') and perception-level (``P'') for \benchmarkname, prompt-level accuracy (``Prompt.") and Inst-level accuracy (``Inst.") for IFEval, and the averaged results across all three benchmarks in the rightmost column.}
    \vspace{-6pt}
    \setlength{\tabcolsep}{4pt}
    \scalebox{.95}{
    \begin{tabular}{lc|>{\centering\arraybackslash}p{1cm}>{\centering\arraybackslash}p{1cm}>{\centering\arraybackslash}p{1.4cm}|c|>{\centering\arraybackslash}p{1.4cm}>{\centering\arraybackslash}p{1cm}>{\centering\arraybackslash}p{1.4cm}|c}
    \toprule
    \multirow{2}{*}{Model} & 
    \multirow{2}{*}{Parameter} & 
    \multicolumn{3}{c|}{MM-IFEval\textsuperscript{M} (ours)} &
    \multirow{2}{*}{MIA\textsuperscript{M}} & 
    \multicolumn{3}{c|}{IFEval\textsuperscript{T}} & 
    \multirow{2}{*}{Avg.} \\
    & & C & P & Avg. & & Prompt. & Inst. & Avg. \\
    \hline
    LLaVA-NeXT-7B \cite{liu2024llavanext} & 7B & 36.8 & 16.0 & 31.6 & 73.2 & 32.0 & 43.3 & 37.7 & 47.5 \\
    LLaVA-OneVision-Qwen2-7B-OV \cite{li2024llava} & 8B & 37.4 & 24.0 & 34.0 & 84.5 & 43.3 & 54.8 & 49.0 & 55.8 \\
    InternVL2-8B \cite{chen2024far} & 8B & 45.2 & 32.0 & 41.9 & 86.2 & 44.6 & 57.0 & 50.8 & 59.6 \\
    InternVL2.5-8B \cite{chen2024expanding} & 8B & 49.6 & 36.0 & 46.2 & 88.5 & 52.2 & 62.4 & 57.3 & 64.0 \\
    \hline
    LLaVA-NeXT-Llama3-8B \cite{liu2024llavanext} & 8B & 45.9 & 21.0 & 39.7 & 83.3 & 45.0 & 56.4 & 50.7 & 57.9 \\
    \rowcolor{lightblue}
    w. \datasetname & - & 59.3 
    & 19.0 
    & 49.2 \footnotesize \hgreen{+9.5}
    & 86.5 \footnotesize \hgreen{+3.2}
    & 50.8 
    & 61.8 
    & 56.3 \footnotesize \hgreen{+5.6}
    & 64.0 \footnotesize \hgreen{+6.1}
    \\
    \rowcolor{bisque}
    w. \dpodatasetname & - & 58.7 
    & 21.0 
    & 49.3 \footnotesize \hgreen{+9.6}
    & 90.0 \footnotesize \hgreen{+6.7}
    & 64.5 
    & 73.7 
    & 69.1 \footnotesize \hgreen{+18.4}
    & \textbf{69.5} \footnotesize \hgreen{+11.6} \\    
    \hline
    Qwen2-VL-7B-Instruct \cite{wang2024qwen2} & 8B & 42.7 & 40.0 & 42.0 & 80.5 & 42.4 & 52.5 & 47.4 & 56.6 \\
    \rowcolor{lightblue}
    w. \datasetname & - & 57.0 
    & 38.0 
    & 52.3 \footnotesize \hgreen{+10.3}
    & 87.7 \footnotesize \hgreen{+7.2}
    & 46.8 
    & 58.4 
    & 52.6 \footnotesize \hgreen{+5.2}
    & 64.2 \footnotesize \hgreen{+7.6} \\
    \rowcolor{bisque}
    w. \dpodatasetname & - & 55.2 
    & 43.0 \footnotesize 
    & 52.2 \footnotesize \hgreen{+10.2}
    & 88.1 \footnotesize \hgreen{+7.6}
    & 55.2 \footnotesize 
    & 64.3 \footnotesize 
    & 59.7 \footnotesize \hgreen{+12.3}
    & \textbf{66.7} \footnotesize \hgreen{+10.1} \\    
    \bottomrule
    \end{tabular}
    }
    \label{tab:main_if}
\end{table*}

\begin{table*}[t]
    \belowrulesep=0pt
    \aboverulesep=0pt
    \centering
    \small
    \caption{\textbf{Main results on VQA benchmarks}, including general knowledge (MMMU \cite{yue2024mmmu}, MMBench \cite{liu2024mmbench}, MMStar \cite{chen2024we}, MMT-Bench \cite{mmtbench}), document understanding (AI2D \cite{kembhavi2016diagram}, OCRBench \cite{liu2024ocrbench}), Chat (MMVet \cite{yu2023mm}) and Hallusion (POPE \cite{li2023evaluatingobjecthallucinationlarge}). Fine-tuning models on \dpodatasetname achieve comparable performance across these benchmarks.}
    \vspace{-6pt}
    \setlength{\tabcolsep}{4pt}
    \scalebox{.99}{
    \begin{tabular}{l|cccc|cc|c|c|c}
    \toprule
    \multirow{2}{*}{Model} &
    \multicolumn{4}{c|}{General} & \multicolumn{2}{c|}{Document} & Chat & Hallusion & \multirow{2}{*}{Avg.} \\
    ~ & MMMU$_\text{val}$ & MMBench$_\text{dev}$ & MMStar & MMT-Bench$_\text{val}$ & AI2D & OCRBench & MMVet & POPE \\
    \midrule
    LLaVA-NeXT-Llama3-8B \cite{liu2024llavanext} & 43.7 & 72.5 & 43.6 & 53.1 & 73.1 & 55.0 & 43.3 & 87.2 & 58.9 \\
    \rowcolor{lightblue}
    w. \datasetname & 45.8 & 69.3 & 44.2 & 53.3 & 71.2 & 55.3 & 46.3 & 88.8 & 59.3 \\
    \rowcolor{bisque}
    w. \dpodatasetname & 44.1 & 72.1 & 43.7 & 53.1 & 72.3 & 56.7 & 43.9 & 86.8 & 59.1 \\
    \hline
    Qwen2-VL-7B-Instruct \cite{wang2024qwen2} & 53.9 & 81.0 & 60.8 & 63.2 & 82.9 & 86.7 & 63.3 & 86.3 & 72.3 \\
    \rowcolor{lightblue}
    w. \datasetname & 54.0 & 79.3 & 57.1 & 61.0 & 81.6 & 81.8 & 61.6 & 89.2 & 70.7 \\
    \rowcolor{bisque}
    w. \dpodatasetname & 54.0 & 81.3 & 58.5 & 63.7 & 83.3 & 86.8 & 66.1 & 85.7 & 72.4 \\
    \bottomrule
    \end{tabular}%
    }
    \label{tab:qwen_table_v3}
\end{table*}

\subsection{Hybrid Evaluation}
Current multi-modal instruction following benchmarks often rely solely on GPT-4o for evaluation.
However, accurately assessing certain constraints, such as numerical conditions (e.g., `\texttt{output in 200 words}', `\texttt{Answer in 5 paragraphs}', `\texttt{Use the word `cat' in the answer twice}'), remains challenging even for GPT-4o.
In contrast, verifiable functions like string matching offer greater precision than judge models for such constraints.
To address this, we propose a hybrid evaluation strategy (see \cref{fig:pipeline}(c)) that employs three methods, including both rule-based Verification and judge models for more robust and precise evaluation.

\noindent (1) \textbf{Rule-based Verification.} For constraints that adhere to a fixed format and involve specific content that can be objectively verified—yet remain challenging for an LLM to assess accurately—we employ a rule-based approach. Specifically, we design a set of predefined functions for different constraint types. The LLM is first prompted to extract the relevant parameters, denoted as \textit{Params}, from the constraint description. When evaluating a constraint that falls within the scope of our rule-based framework, we use \textit{Params} and the model’s output as inputs to the predefined function to determine compliance.

\noindent (2) \textbf{LLM-based Direct Judgment.} This method is primarily used for evaluating constraints that can be easily and unambiguously verified based on the model's output. It is applicable to constraints where correctness is straightforward to determine, such as those requiring the inclusion of specific words or phrases. For instance, a constraint like “Use the word ‘inspiration’ or its synonyms at least twice in the response” does not follow a strict format and cannot be assessed using a rule-based approach. Instead, we directly leverage an LLM to determine whether the constraint is satisfied.

\noindent (3) \textbf{LLM-based Comparative Judgment.} Some constraints, particularly those related to tone, style, or role-playing, are difficult to evaluate directly. To improve judgment accuracy, we adopt a comparative approach. Specifically, we generate a second model output using a nearly identical prompt but without the constraint under evaluation. The LLM-based evaluator is then provided with both outputs and asked to compare them, determining whether the model’s response with the constraint in the prompt adheres more closely to the expected requirement.

\section{Experiments}

\noindent \textbf{Benchmarks.} We select the following benchmarks to demonstrate that models fine-tuned on \datasetname and \dpodatasetname enhance instruction following without compromising performance on other VQA tasks:
(1) \textbf{Instruction Following benchmarks}, including MIA-Bench \cite{qian2024mia}, IFEval \cite{zhou2023instruction}, and our proposed \benchmarkname. To be noted, IFEval is a language-only benchmark while others are both multi-modal benchmarks.
(2) \textbf{VQA Benchmarks}, including MMMU \cite{yue2024mmmu}, MMBench \cite{liu2024mmbench}, MMStar \cite{chen2024we}, AI2D \cite{kembhavi2016diagram}, OCRBench \cite{liu2024ocrbench}, MMVet \cite{yu2023mm}, POPE \cite{li2023evaluatingobjecthallucinationlarge} and MMT-Bench \cite{mmtbench}.

\noindent \textbf{Implementation Details.}
We conducted SFT and DPO fine-tuning experiments on two representative MLLMs: Qwen2-VL-7B-Instruct \cite{wang2024qwen2} and LLaVA-Next-Llama3-8B \cite{liu2024llavanext}, using our custom datasets \datasetname for supervised fine-tuning (SFT) and \dpodatasetname for direct preference optimization (DPO).
For the SFT phase, we used a batch size of 128 and a learning rate of 1e-5. For the DPO phase, we used a learning rate of 5e-7 with the batch size of 16. We implemented our training pipeline with the help of LLaMA-Factory and evaluation pipeline under VLMEvalkit \cite{duan2024vlmevalkit}.

\subsection{Results about \datasetname and \dpodatasetname}
\noindent \textbf{Consistently Improvements on Instruction Following Benchmarks.}
As shown in \cref{tab:main_if}, both \datasetname and \dpodatasetname significantly enhance the model's performance in instruction following benchmarks.
Fine-tuning LLaVA-Next and Qwen2-VL on \datasetname yielded significant averaging performance gains of 6.1$\%$ and 7.6$\%$ points, respectively.
Furthermore, applying DPO with \dpodatasetname also led to notable improvements for LLaVA-Next and Qwen2-VL, with average gains of 11.6$\%$ and 10.1$\%$ points.
Such improvements demonstrate the effectiveness of \pipelinename in constructing high-quality training data.

\noindent \textbf{Comparable Results on VQA Benchmarks.}
To show that fine-tuning on \datasetname and \dpodatasetname improves instruction following without degrading performance on other VQA tasks, we analyzed model performance on other widely used benchmarks, as detailed in \cref{tab:qwen_table_v3}.
Results indicate that models fine-tuning with \datasetname and \dpodatasetname demonstrate comparable performance across these benchmarks.

\noindent \textbf{SFT vs DPO.}
As evidenced by \cref{tab:main_if} and \cref{tab:qwen_table_v3}, DPO using \dpodatasetname significantly surpasses SFT on \datasetname.
This is likely due to negative samples of DPO, which are essential for training models to respect constraints, particularly in our data with multiple and diverse constraints.
Additionally, the Kullback–Leibler (KL) divergence in DPO preserves the model's generalization, as demonstrated in \cref{tab:qwen_table_v3}.


\subsection{Leaderboard of \benchmarkname}
We present the performance comparison results of various MLLMs on our \benchmarkname in \cref{tab:leaderboard}, including both proprietary MLLMs such as GPT-4o \cite{hurst2024gpt} and Claude-3.5 \cite{claude} and open-source MLLMs such as LLaVA-Next \cite{liu2024llavanext}, LLaVA-OneVision \cite{li2024llava}, InternVL \cite{chen2024far,chen2024expanding}, and Qwen2-VL \cite{wang2024qwen2}.

\noindent \textbf{\benchmarkname is Challenging.} Results on \cref{tab:leaderboard} demonstrate that multimodal instruction following is still a challenging and unsolved task for current MLLMs, specifically for the perception-level problems.
The propriety models GPT-4o and Claude-3.5V-Sonnet establish top-tier average performance with scores of 64.6 and 61.7, respectively.
The leading open-source MLLM, Qwen2-VL-72B merely achieves an overall accuracy of 50.8.
We attribute the performance gap between proprietary and open-source models to the scarcity of high-quality open-source training data for instruction following.
As a result of our \dpodatasetname, Qwen2-VL-7B fine-tuned via our optimized DPO approach achieves a score of 52.2, demonstrating a 24.3\% relative improvement over its baseline (42.0), and even surpasses the larger Qwen2VL-72B model.
We hope our \benchmarkname benchmark motivates further exploration into improving MLLM instruction-following.

\noindent \textbf{Benchmark Examples.} Please refer to the Appendix for visual examples of \benchmarkname, including images and instructions with constraints for both compose-level and perception-level problems.

\definecolor{lightgray}{gray}{0.9}  

\begin{table}[t]
\centering
\small
\caption{\textbf{Evaluation of various MLLMs on \benchmarkname}. We report the accuracy of easy and difficult problems and the average accuracy across all problems. The C-Level and P-Level refer to the compose-level and perception-level problems, respectively. The \textbf{best} performance in each section is highlighted in \textbf{bold}.}
\vspace{-6pt}
\resizebox{.48\textwidth}{!}{%
\begin{tabular}{lc ccc}
\toprule
Model & Param & C-Level & P-Level & Avg. \\
\midrule
\multicolumn{5}{l}{\textit{Proprietary MLLMs}} \\ 
\midrule
Claude-3.5V-Sonnet \cite{claude} & - & 67.5 & \textbf{44.0} & 61.7 \\
GPT-4o-mini \cite{hurst2024gpt}      & - &  70.4 & 40.0 & 62.8 \\
GPT-4o (20240806) \cite{hurst2024gpt} & - & \textbf{71.5} & \textbf{44.0} & \textbf{64.6} \\
\midrule
\multicolumn{5}{l}{\textit{Open-Source MLLMs}} \\
\midrule
LLaVA-NeXT-7B \cite{liu2024llavanext} & 7B & 36.8 & 16.0 & 31.6 \\
LLaVA-OneVision-Qwen2-7b-OV \cite{li2024llava} & 8B &  37.4 & 24.0 & 34.0 \\
MiniCPM-V-2.6 \cite{yao2024minicpm} & 8B &  39.2 & 32.0 & 37.4 \\
InternVL2-8B \cite{chen2024far} & 8B & 45.2 & 32.0 & 41.9 \\
InternVL2-40B \cite{chen2024far} & 40B & 48.0 & 36.0 & 45.0 \\
InternVL2.5-8B \cite{chen2024expanding} & 8B &  49.6 & 36.0 & 46.2 \\
InternVL2.5-26B \cite{chen2024expanding} & 8B &  53.5 & 32.0 & 48.1 \\
Qwen2-VL-72B-Instruct \cite{wang2024qwen2} & 72B &  53.4 & \textbf{43.0} & 50.8 \\


\midrule
LLaVA-NeXT-Llama3-8B \cite{liu2024llavanext} & 8B & 45.9 & 21.0 & 39.7 \\
\rowcolor{bisque}
+ \dpodatasetname    & - & \textbf{58.7} & 21.0 & 49.3 \\
\midrule
Qwen2-VL-7B-Instruct \cite{wang2024qwen2} & 8B & 42.7 & 40.0 & 42.0 \\
\rowcolor{bisque}
+ \dpodatasetname    & - &  55.2 & \textbf{43.0} & \textbf{52.2} \\
\bottomrule
\end{tabular}
\vspace{-6pt}
}
\label{tab:leaderboard}
\end{table}
\begin{table}[t]
    \centering
    \small
    \caption{
    \textbf{Ablation studies across different DPO settings}, including randomly deleting constraints (second row to fourth row) or prompting MLLMs without images (bottom row) to generate negative responses. Avg. refers to the average score of three IF benchmarks.}
    \vspace{-6pt}
    \resizebox{.48\textwidth}{!}{ 
    \begin{tabular}{l|ccc|c}
    \toprule
    Model & 
    MM-IFEval &
    MIA & 
    IFEval &
    Avg. \\
    \midrule
    Qwen2-VL-7B-Instruct & 42.0 & 80.5 & 47.4 & 56.6 \\
    {\raggedright + DPO (-33\% cons)} & 51.5 & \textbf{88.2} & 57.9 & 65.8  \\
    {\raggedright + DPO (-66\% cons)} & 51.2 & 88.0 & 58.4 & 65.9 \\
    {\raggedright + DPO (-100\% cons)} & \textbf{52.2} & 88.1 & \textbf{59.7} & \textbf{66.7} \\
    {\raggedright + DPO (w/o img)} & 48.4 & 86.9 & 54.7 & 63.4 \\
    \midrule
    LLaVA-NeXT-Llama3-8B & 39.7 & 83.3 & 50.7 & 57.9 \\
    {\raggedright + DPO (-33\% cons)} & \textbf{50.4} & 87.2 & 64.3 & 67.3  \\
    {\raggedright + DPO (-66\% cons)} & 48.7 & 86.8 & \textbf{69.7} & 68.4 \\
    {\raggedright + DPO (-100\% cons)} & 49.3 & \textbf{90.0} & 69.1 & \textbf{69.5} \\
    {\raggedright + DPO (w/o img)} & 44.7 & 85.9 & 64.8 & 65.2 \\
    \bottomrule
    \end{tabular}
    }
    \label{tab:dpo_ablation}
    \vspace{-6pt}
\end{table}

\subsection{Ablation Studies}

\noindent \textbf{Ablation Studies on Different DPO Settings.}
In \cref{tab:dpo_ablation}, we present an ablation study on various strategies for constructing pairwise preference data for Direct Preference Optimization (DPO). These strategies primarily include: (1) generating rejected responses by randomly removing constraints from the instruction (second to fourth rows), and (2) prompting MLLMs without providing image inputs to generate rejected responses (bottom row).

We conduct experiments on both the Qwen2-VL-7B-Instruct and LLaVA-NeXT-Llama3-8B models. As shown in \cref{tab:dpo_ablation}, all DPO variants exhibit strong robustness, consistently outperforming the baseline. Among the four evaluated strategies, removing 100\% of the constraints to generate rejected responses achieves the best performance, whereas omitting image inputs yields the weakest performance. Furthermore, we observe a consistent trend: as the proportion of removed constraints increases from 33\% to 100\%, the performance of the resulting DPO models improves accordingly. This suggests that removing more constraints amplifies the semantic gap between preferred and rejected responses, thereby enhancing the effectiveness of contrastive learning during DPO training.

Based on these findings, we adopt the 100\%-constraint removal strategy as the default approach for constructing the DPO data in \dpodatasetname.

\section{Conclusion}
This paper contributes to the field of multimodal instruction-following by exploring pipelines for training data collection and proposing a challenging benchmark.
We present \pipelinename, a pipeline designed to generate image-instruction pairs, subsequently used to construct \datasetname for SFT and \dpodatasetname for DPO.
We also analyze the limitations of existing multimodal instruction following benchmarks and propose \benchmarkname, a benchmark featuring diverse instruction types and a hybrid evaluation strategy that combines rule-based methods with an LLM-based judge.
We hope this work inspires further research into improving the instruction-following ability of Multimodal Large Language Models, a critical step towards realizing their potential in diverse and impactful applications.

{
    \small
    \bibliographystyle{ieeenat_fullname}
    \bibliography{main}
}

\newpage
\appendix
\clearpage
\setcounter{page}{1}
\maketitlesupplementary
\section{MM-IFEval}

\subsection{An overview of Constraints and Instructions}

\subsubsection{Constraints}
\label{app:constraints_category}

\begin{figure*}[t]
    \centering
    \includegraphics[width=.8\linewidth]
    {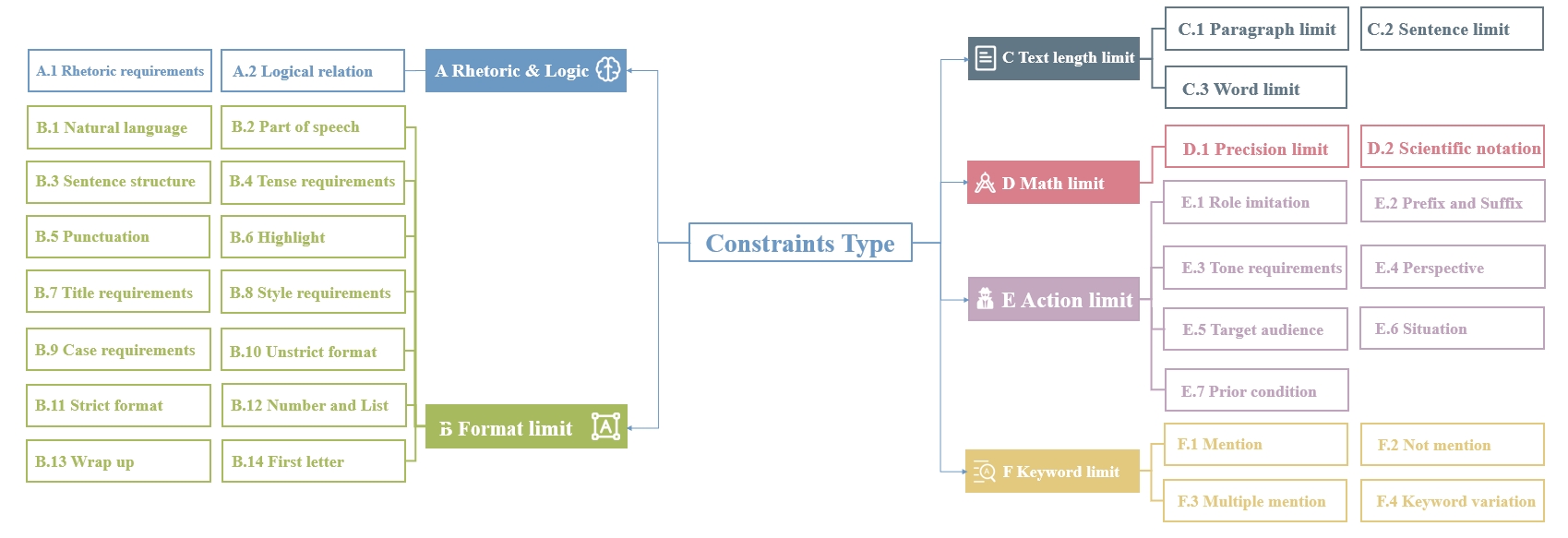}
    \caption{\textbf{Demonstration of constraints categories}. We designed 6 main categories for all the constraints used, with a total of 32 subcategories}
    \label{fig:Taxonomy_of_Constraints}
    \vspace{-6pt}
\end{figure*}

Based on daily use cases and existing research, we have identified six main categories of constraints, which can be further divided into 32 specific constraint types shown in Fig.~\ref{fig:Taxonomy_of_Constraints}. In this section, we introduce and exemplify these six major constraint categories. For detailed descriptions and examples of all 32 subcategories, please refer to Table~\ref{tab:subconstraints}.

\noindent\textbf{Text Length Requirements}. In this category, we focus on the length of the response, including the number of paragraphs, sentences, and words. We also consider the length of the response in the aspect of poetry or ``Use yes or no to answer the question". It must be noted that we do not require the model to follow the strict requirement in exact numbers like ``\textit{The response must be exactly 56 words}". The constraints we propose in this category are based on reality, with precise numerical requirements only at the sentence or paragraph level, and of moderate size; the rest of the constraints are used to limit by ranges like ``\textit{The response must be between 100 and 150 words}", which aligns with the task that people tend to encounter in real-world scenarios.

\noindent\textbf{Mathematical Requirements}. This category includes constraints related to the most common part of answering mathematical problems like precision, scientific notation, and other mathematical requirements. For example, ``\textit{Keep two decimal places for the number in the answer}", ``\textit{Please round up all the numbers in the answer}", or ``\textit{Don't include specific numbers in your answers. Compare numbers with their relative sizes}".

\noindent\textbf{Language \& Formatting Requirements}. This category includes constraints related to the language and formatting of the response, such as answering in a specific language, using a specific format like JSON, or using a specific style like poetry. Requirements for tense, writing style, numbering, list, and other language-related or formatting-related aspects are also included in this category.

\noindent\textbf{Rhetoric \& Logic Requirements}. ``Rhetoric" refers to the art of using language to persuade or influence, while ``Logic" refers to the principles of reasoning and argumentation. This category includes constraints related to the rhetoric and logic of the response, such as the use of metaphor, simile, cause-and-effect relationship, conditional statement, and other rhetoric and logic-related aspects.

\noindent\textbf{Action Requirements}. ``Action" refers to the action that the model should take like a human. We define this category as the constraints that require the model to perform a specific action, such as tone, role imitation, use specific prefix or suffix, or acting like under some specific situation. We hope this category can help us to evaluate the ability of the model to follow instructions and perform actions in more complex and realistic scenarios.

\noindent\textbf{Keyword Requirements}. ``Keyword" refers to the specific words or phrases that the model should include or avoid in the response. This category includes constraints related to the response keyword, such as the use of specific keywords, the avoidance of specific keywords, or the variation of specific keywords. For example, ``Use at least three synonyms for `innovation,' such as `breakthrough,' `new approach,' or `invention,' spread throughout your text."

\subsubsection{Instruction Tasks}
\label{app:task_pool}
For source datasets lacking original task instructions, we constructed a diverse task pool containing 18 instructions that encourage open-ended responses from models. These instructions can be categorized into five task types: Descriptive Analysis, Emotional \& Perspective, Creative Writing, Social Media \& Content, and Roleplay. The classification information and examples of the instructions are shown in Table~\ref{tab:task_pool}.

\subsection{Perception-level Problems}
\begin{figure}[H]
    \centering
    \includegraphics[width=.4\linewidth]{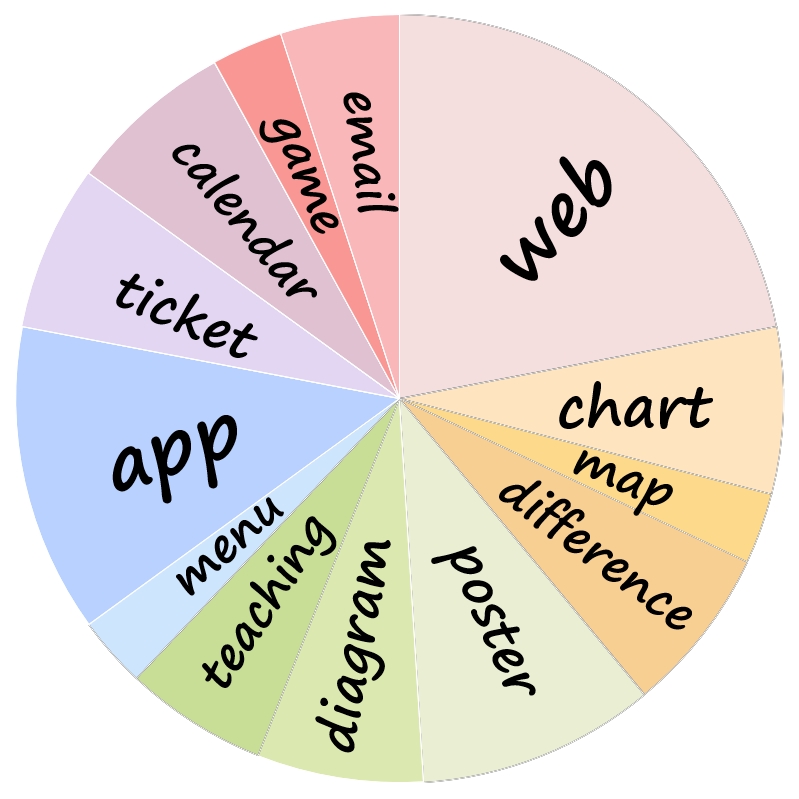}
    \caption{\textbf{Image Source Distribution in perception-level problems}.Perception-level problems in MM-IFEval presents a systematic categorization of 100 challenging vision-based instruction-following tasks, organized into 13 distinct classes according to image content characteristics and task complexity.}
    \label{fig:bench_vision_part_category}
    \vspace{-10pt}
\end{figure}
\begin{figure*}[htbp]
    \centering
    \includegraphics[width=\linewidth]{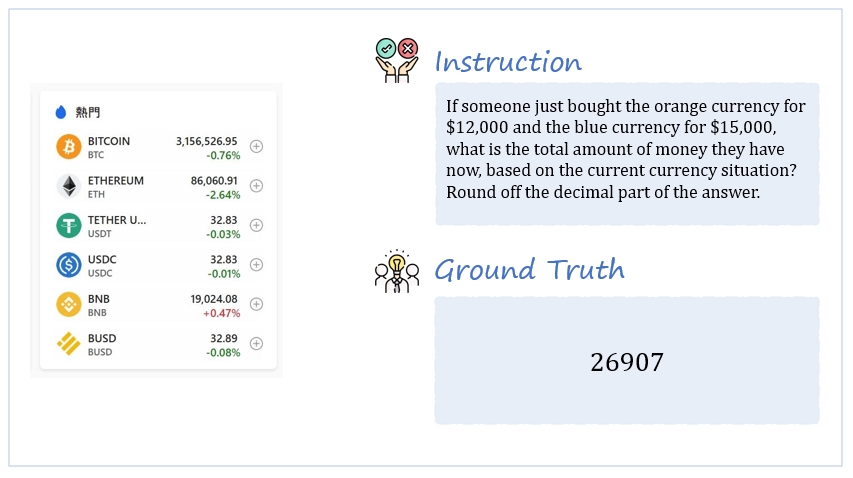}
    \caption{\textbf{A perception-level problem example from the MM-IFEval benchmark in the web category. }}
    \label{fig:v_web}
    \vspace{-10pt}
\end{figure*}
\begin{figure*}[htbp]
    \centering
    \includegraphics[width=\linewidth]{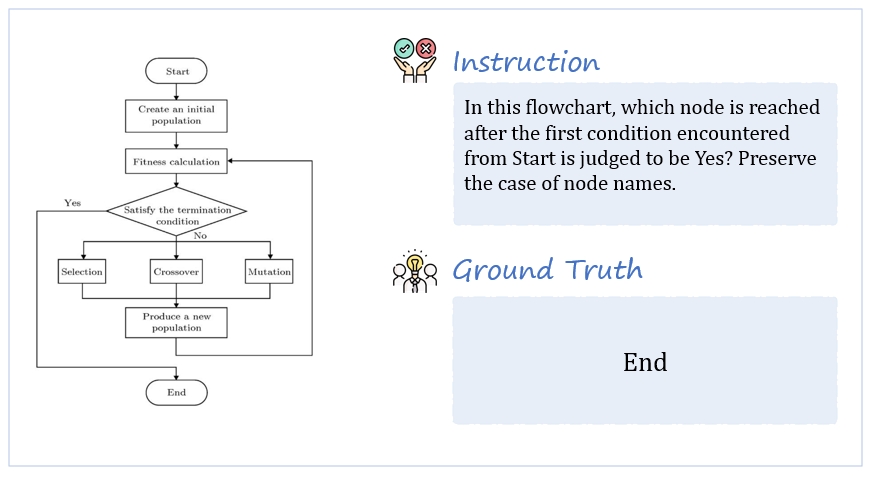}
    \caption{\textbf{A perception-level problem example from the MM-IFEval benchmark in the diagram category. }}
    \label{fig:v_diagram}
    \vspace{-10pt}
\end{figure*}
\begin{figure*}[htbp]
    \centering
    \includegraphics[width=\linewidth]{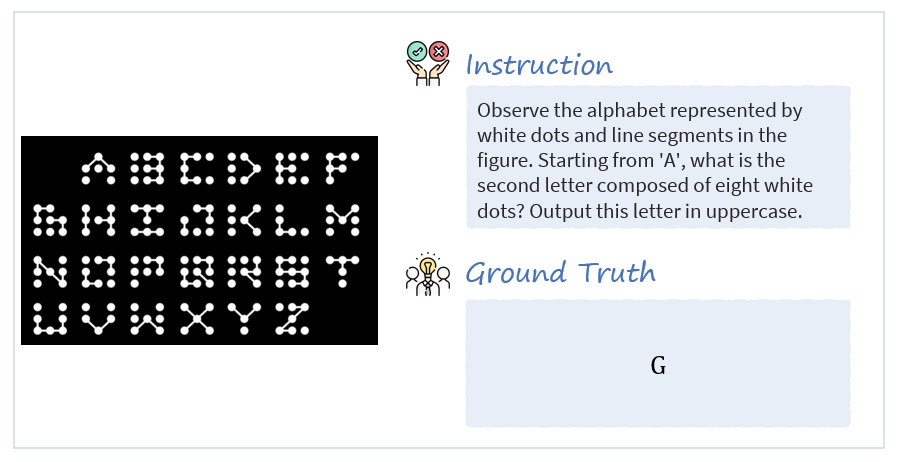}
    \caption{\textbf{A perception-level problem example from the MM-IFEval benchmark in the poster category. }}
    \label{fig:v_poster}
    \vspace{-10pt}
\end{figure*}
\begin{figure*}[htbp]
    \centering
    \includegraphics[width=\linewidth]{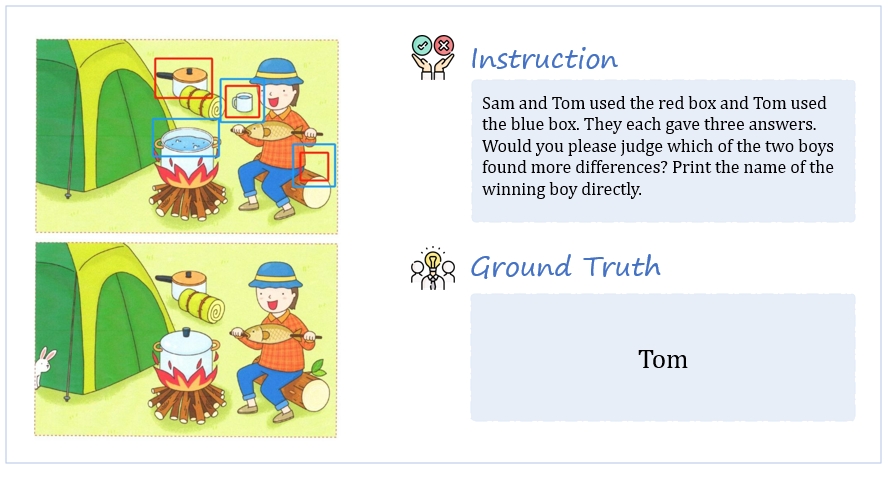}
    \caption{\textbf{A perception-level problem example from the MM-IFEval benchmark in the finding difference category. }}
    \label{fig:v_difference}
    \vspace{-10pt}
\end{figure*}

Perception-level problems in MM-IFEval comprise 100 carefully crafted questions with strong image-constraint correlations. The images can be categorized into 13 information-rich and complex domains shown in Figure~\ref{fig:bench_vision_part_category}. Figures~\ref{fig:v_web}, \ref{fig:v_diagram}, \ref{fig:v_poster}, and \ref{fig:v_difference} present representative examples from the web interface, diagram, poster, and visual difference categories, respectively, demonstrating the diverse visual challenges incorporated in our benchmark.

\section{Image Sources}
The quality of the image source is crucial for the performance of the model. Except of this, the diversity of the image source is also important to fully utilize or evaluate the ability of the model. We use the following image source:
\begin{itemize}
    \item \textbf{Natural Scene}: The natural scene is the most common image source, which is most used in the real-world like the image of a beautiful landscape, a busy street, or a crowded cafe. In this part, we sample images from CC3M\cite{sharma2018conceptual} and ALLaVA\cite{chen2024allava}.
    \item \textbf{UI Interface}: The UI interface is the image from the UI interface of the website and mobile application. It is crucial because it represents a significant portion of real-world multimodal interactions where users need to understand and interact with digital interfaces. We collected diverse mobile app UI images from the RICO\cite{deka2017rico} dataset and web UI images from the MultiUI\cite{liu2024harnessingwebpageuistextrich} dataset.
    \item \textbf{Diagram \& Chart}: The diagram and chart are the image that contains some specific information like the data, the relationship between the data, or the change of the data. We collect diagram and chart images from ChartQA\cite{masry2022chartqa} dataset, which contains diverse diagram and chart images.
    \item \textbf{Mathematic}: The math problem is the image that contains a math problem, which is a common task in the real-world like the problem of the math, the solution of the math problem, or the calculation of the math problem. We collect math problem images from Geo170k\cite{gao2023g} dataset, which contains diverse geometry problem images.
\end{itemize}

\section{MM-IFEngine Prompt Template}
\begin{figure*}[htbp]
    \centering
    \includegraphics[width=\linewidth]{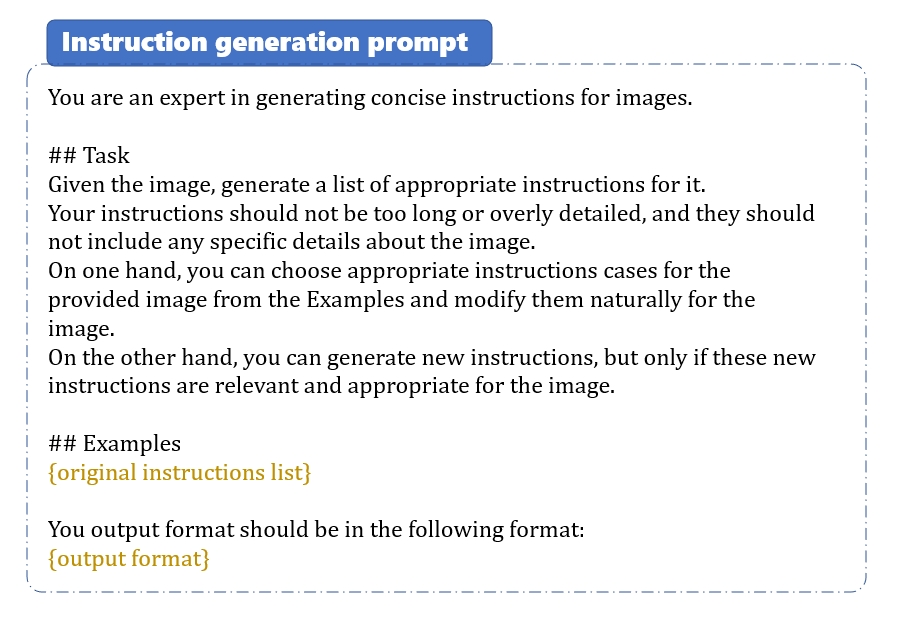}
    \caption{Prompt template for image generation instructions using a large language model in MM-IFEngine.}
    \label{fig:instruction_generation}
    \vspace{-10pt}
\end{figure*}

\begin{figure*}[htbp]
    \centering
    \includegraphics[width=\linewidth]{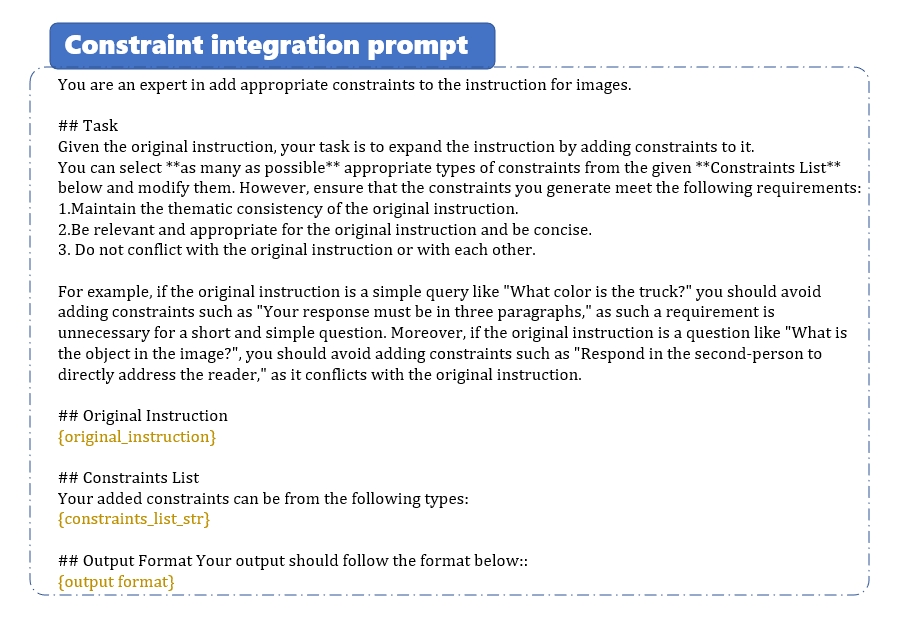}
    \caption{prompt template for integrating constraints in MM-IFEngine.}
    \label{fig:constraint_integration}
    \vspace{-10pt}
\end{figure*}

MM-IFEngine provides a scalable pipeline for mass-producing instruction-following datasets for multimodal large language models, functioning effectively regardless of whether source datasets contain original instructions. This engine enables systematic augmentation of existing visual datasets with diverse instruction-following tasks. Figures~\ref{fig:instruction_generation} and \ref{fig:constraint_integration} demonstrate representative prompt templates from MM-IFEngine's two core components: the instruction generation module and the constraint integration module, respectively, illustrating the methodology behind our automated data construction process.
\section{MM-IFInstruct and MM-IFDPO Dataset}
\begin{figure*}[htbp]
    \centering
    \includegraphics[width=\linewidth]{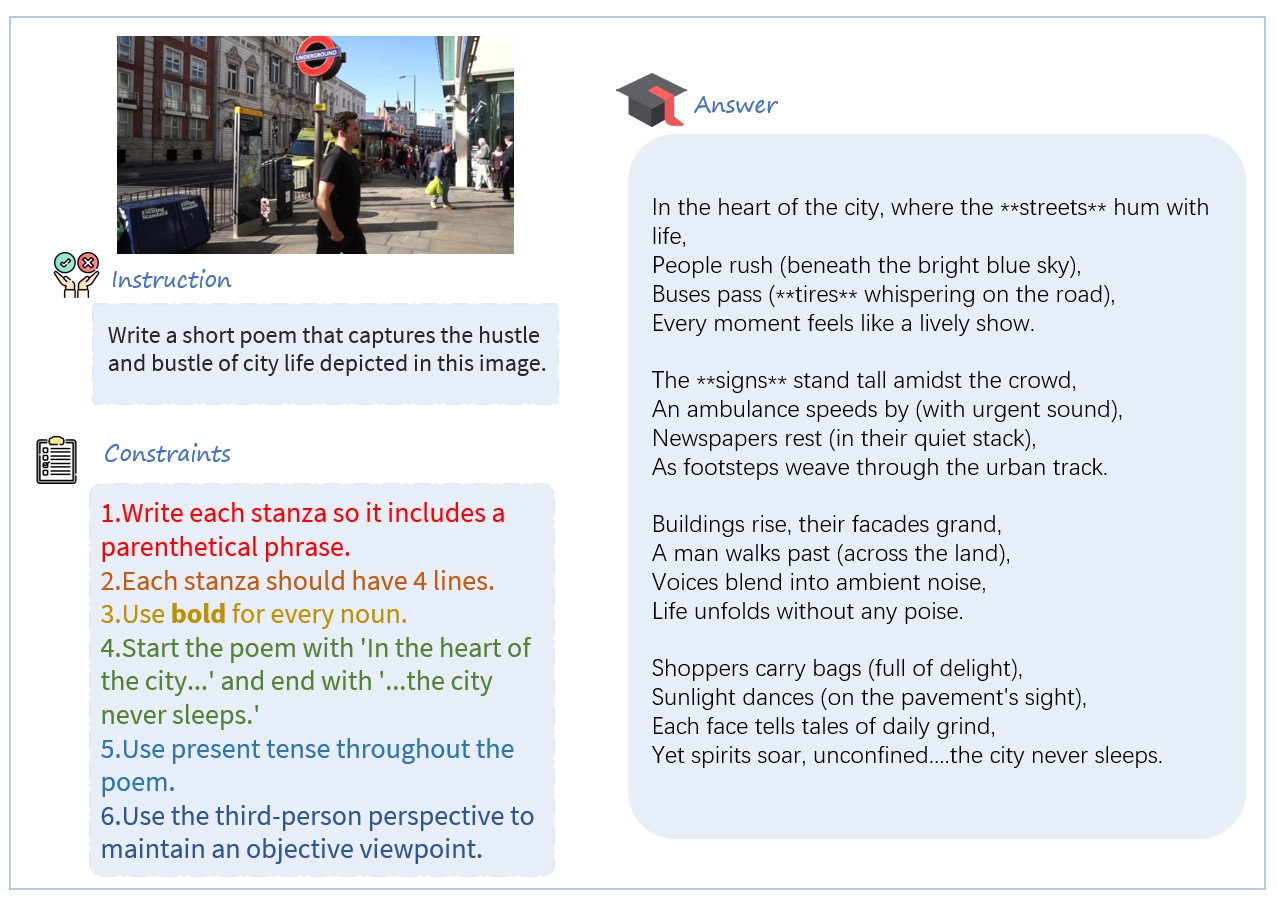}
    \caption{A sample constructed by MM-IFEngine pipeline from cc3m dataset}
    \label{fig:cc3m}
\end{figure*}

\begin{figure*}[htbp]
    \centering
    \includegraphics[width=\linewidth]{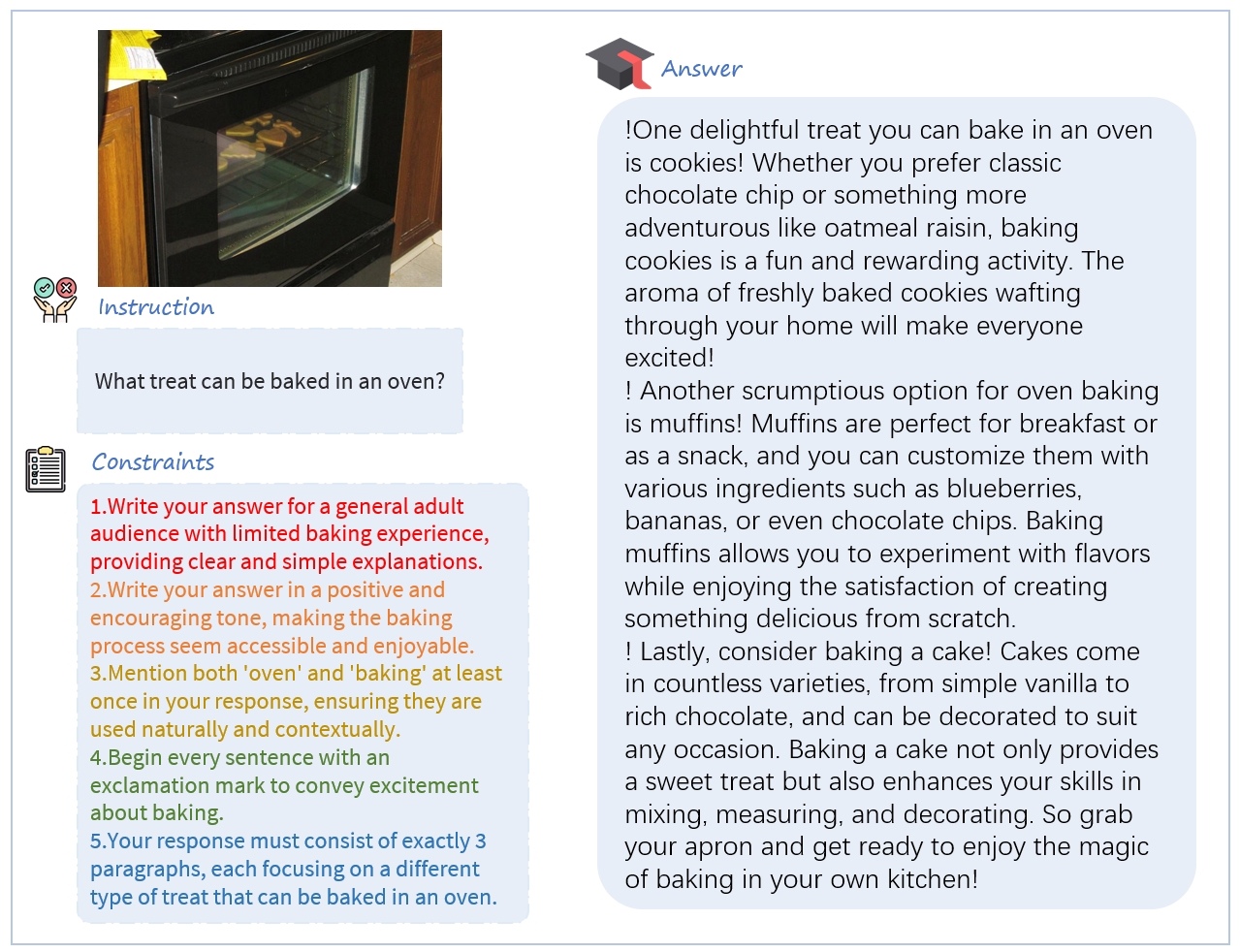}
    \caption{A sample constructed by MM-IFEngine pipeline from Allava dataset}
    \label{fig:allava}
    \vspace{-10pt}
\end{figure*}

\begin{figure*}[htbp]
    \centering
    \includegraphics[width=\linewidth]{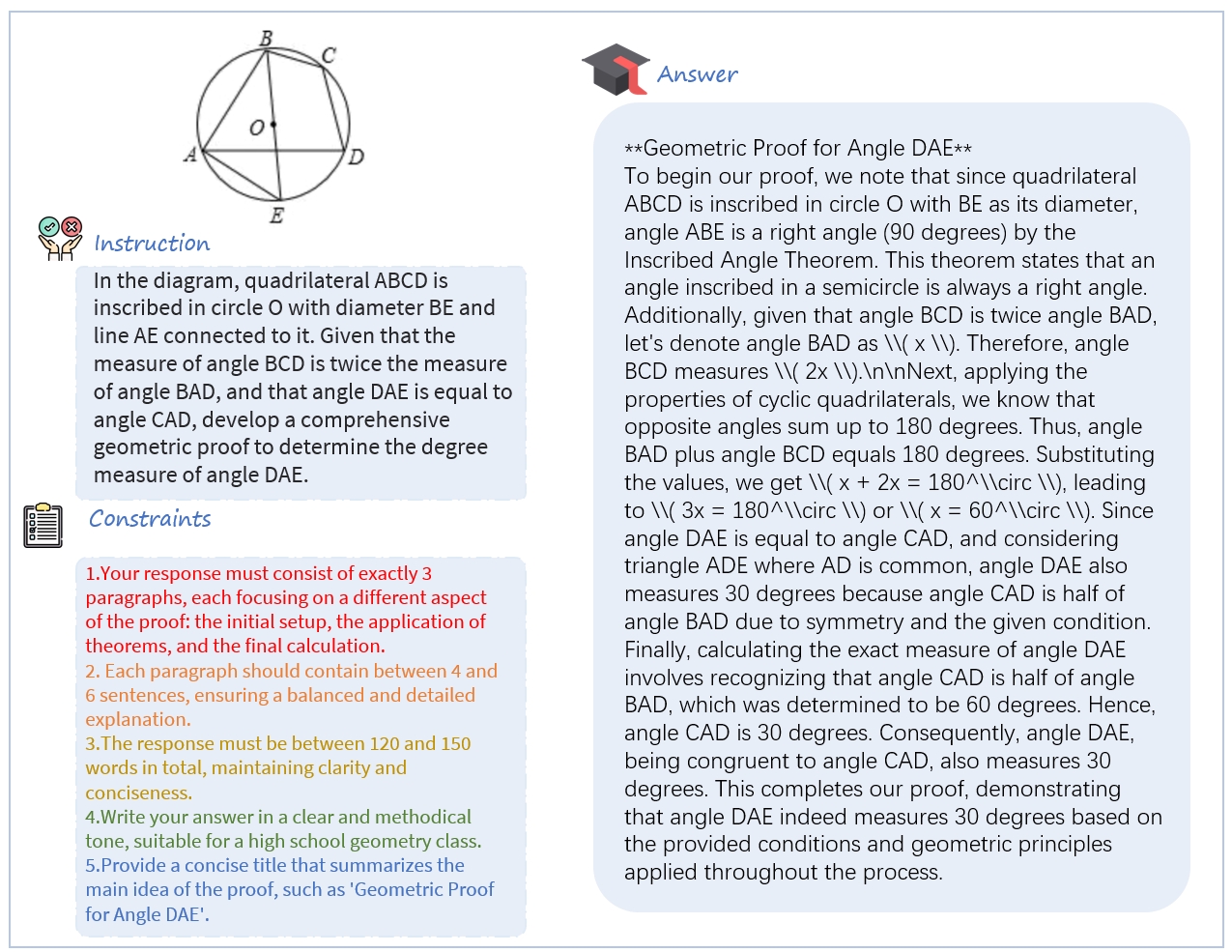}
    \caption{A sample constructed by MM-IFEngine pipeline from geo170k dataset}
    \label{fig:diversity}
    \vspace{-10pt}
\end{figure*}

\begin{figure*}[htbp]
    \centering
    \includegraphics[width=\linewidth]{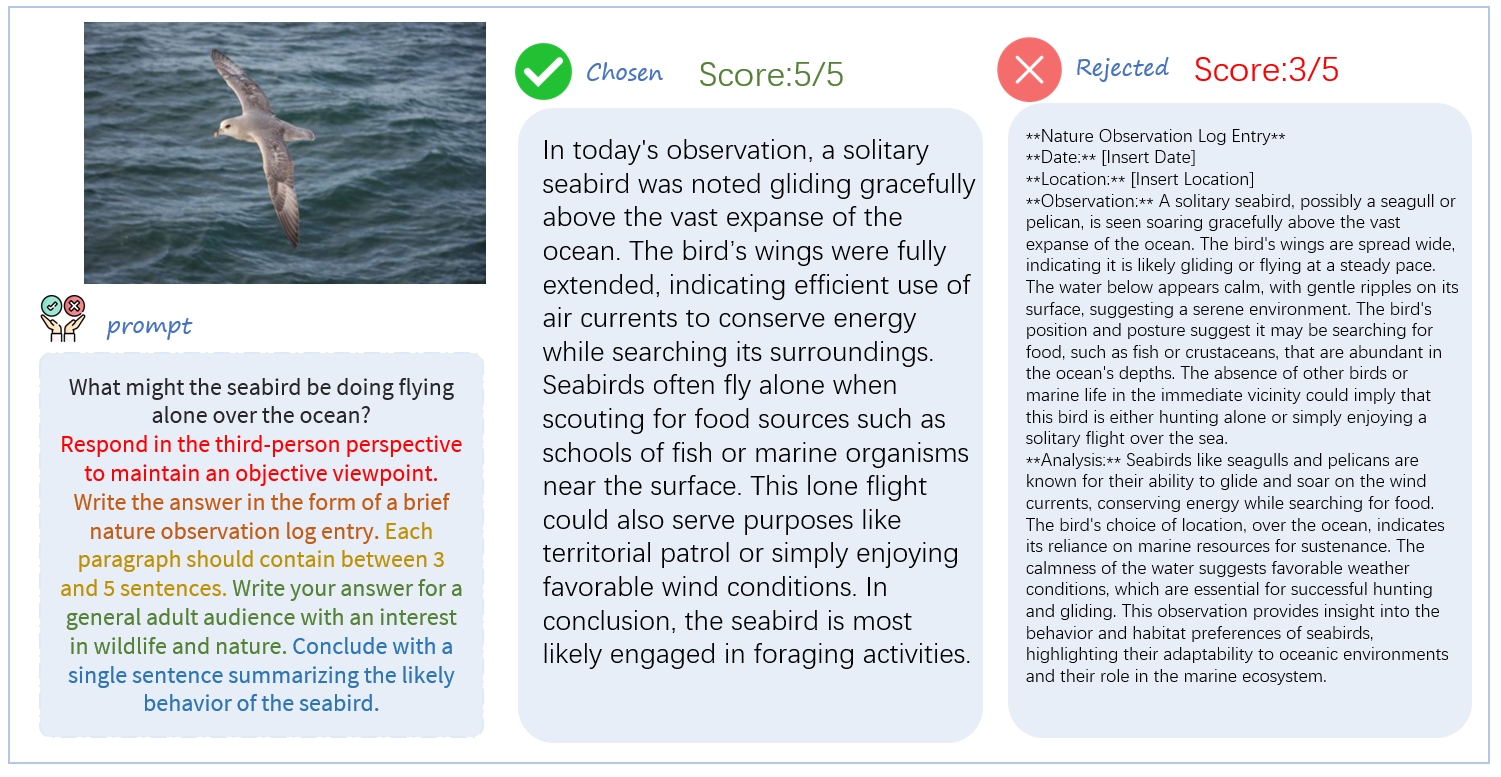}
    \caption{A DPO training set sample, where the rejected data is obtained by removing 33\% of the constraints}
    \label{fig:DPO}
    \vspace{-10pt}
\end{figure*}

Our MM-IFInstruct dataset integrates three distinct data sources: CC3M (without original instructions), ALLaVA (with pre-existing questions), and a diversity collection composed of MultiUI, ChartQA, and Geo170k. To create the MM-IFDPO dataset for preference optimization, we randomly removed 33\% of constraints from the MM-IFInstruct samples to generate rejected examples. Figures~\ref{fig:cc3m}, \ref{fig:allava}, and \ref{fig:diversity} illustrate representative samples derived from CC3M, ALLaVA, and our diversity collection, respectively, while Figure~\ref{fig:DPO} demonstrates an example pair from the MM-IFDPO dataset showing both preferred and rejected instructions.

\section{Evaluation}
\subsection{Rule-based}
\begin{figure*}[htbp]
    \centering
    \includegraphics[width=\linewidth]{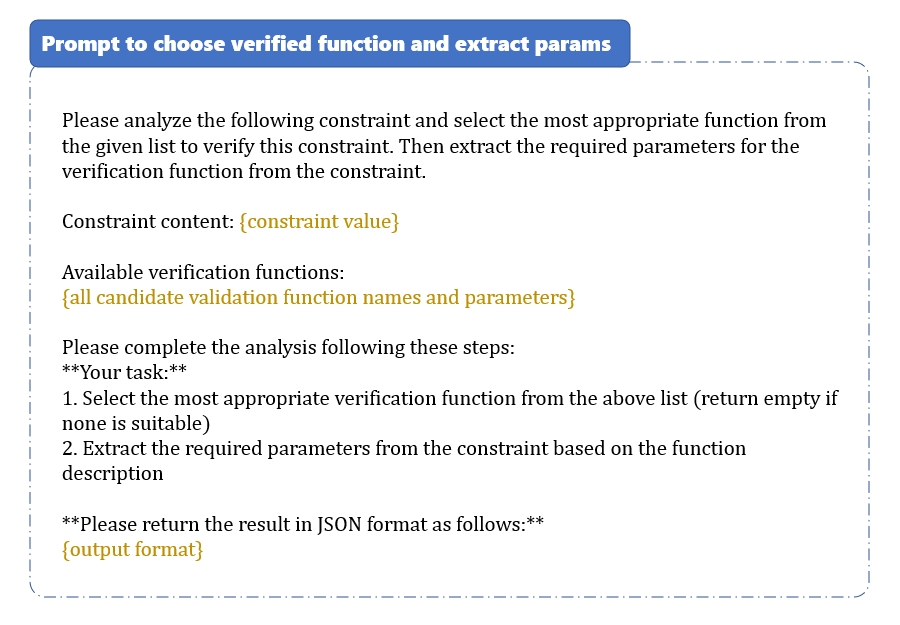}
    \caption{Prompt template for automated verification function selection and paramater extraction}
    \label{fig:choose_f_and_p}
    \vspace{-10pt}
\end{figure*}

We identified 10 constraint subcategories from our taxonomy of 32 that could be algorithmically verified. For these selected constraints, we developed specialized verification functions with targeted parameters. For efficiency, we employed large language models to analyze each constraint specification, select the most appropriate verification function, and extract the necessary parameters. All selections were subsequently validated through manual review to ensure the accuracy and quality of both the function selection and their parameters. The prompt template used for function selection and parameter extraction is illustrated in Figure~\ref{fig:choose_f_and_p}, while Table~\ref{tab:verification_functions} provides a comprehensive overview of all verification functions with their corresponding parameter examples.
\subsection{Compare Judge Method}
\begin{figure*}[htbp]
    \centering
    \includegraphics[width=\linewidth]{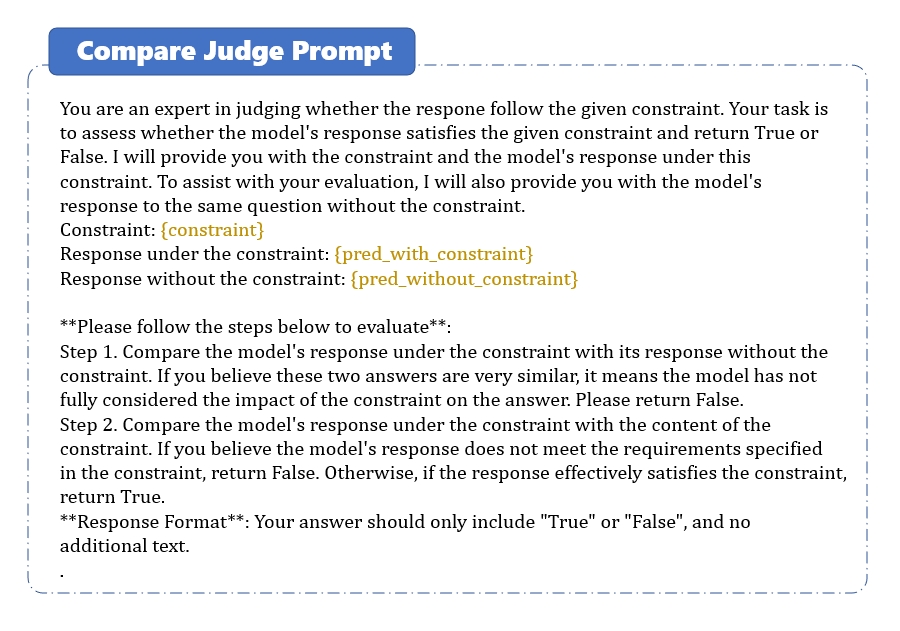}
    \caption{Prompt template for Compare Judge Method}
    \label{fig:compare_judge}
    \vspace{-10pt}
\end{figure*}

Recent works\cite{lu2024wildvision, fang2025creation} have shown that GPT-4o has the ability to compare two responses from models. For constraint types lacking objective evaluation metrics (such as tone requirements or role imitation), we implemented a comparative assessment method. This approach requires the model under evaluation to generate two responses: one adhering to the target constraint and another without the constraint. A judge model then analyzes both outputs to determine whether significant differences exist between them, thereby more accurately assessing whether the model has successfully followed these subjective constraints. Figure~\ref{fig:compare_judge} illustrates the prompt used in this comparative evaluation process. 
\subsection{Direct Judge Method}
\begin{figure*}[htbp]
    \centering
    \includegraphics[width=\linewidth]{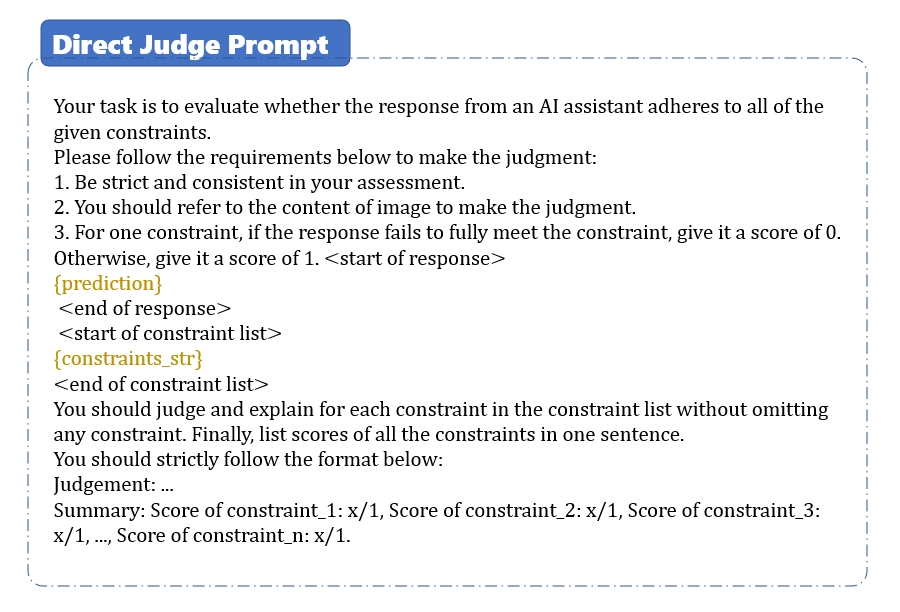}
    \caption{Prompt template for Direct Judge Method}
    \label{fig:direct_judge}
    \vspace{-10pt}
\end{figure*}

The Direct Judge method provides the constraint and answer of the model under test directly to the Judge model, and its prompt template is shown in Figure~\ref{fig:direct_judge}.

\begin{figure*}[htbp]
    \centering
    \includegraphics[width=\linewidth]{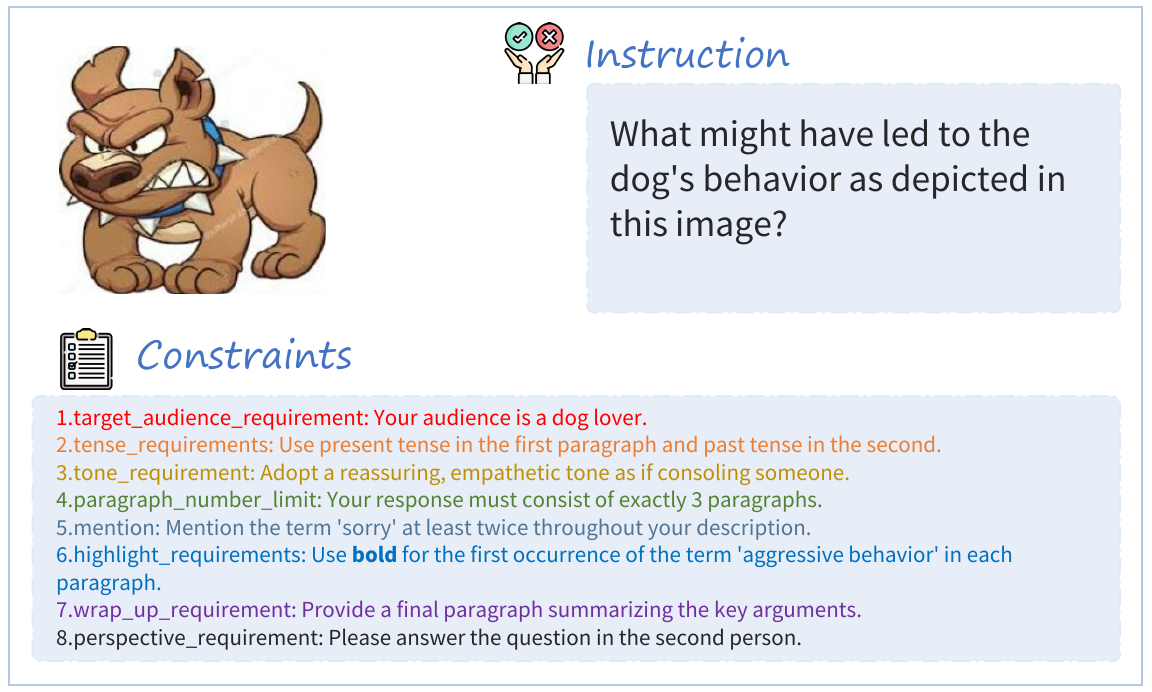}
    \caption{\textbf{A compose-level problem example from the MM-IFEval benchmark in the general image category.}}
    \label{fig:t_general}
    \vspace{-10pt}
\end{figure*}
\begin{figure*}[htbp]
    \centering
    \includegraphics[width=\linewidth]{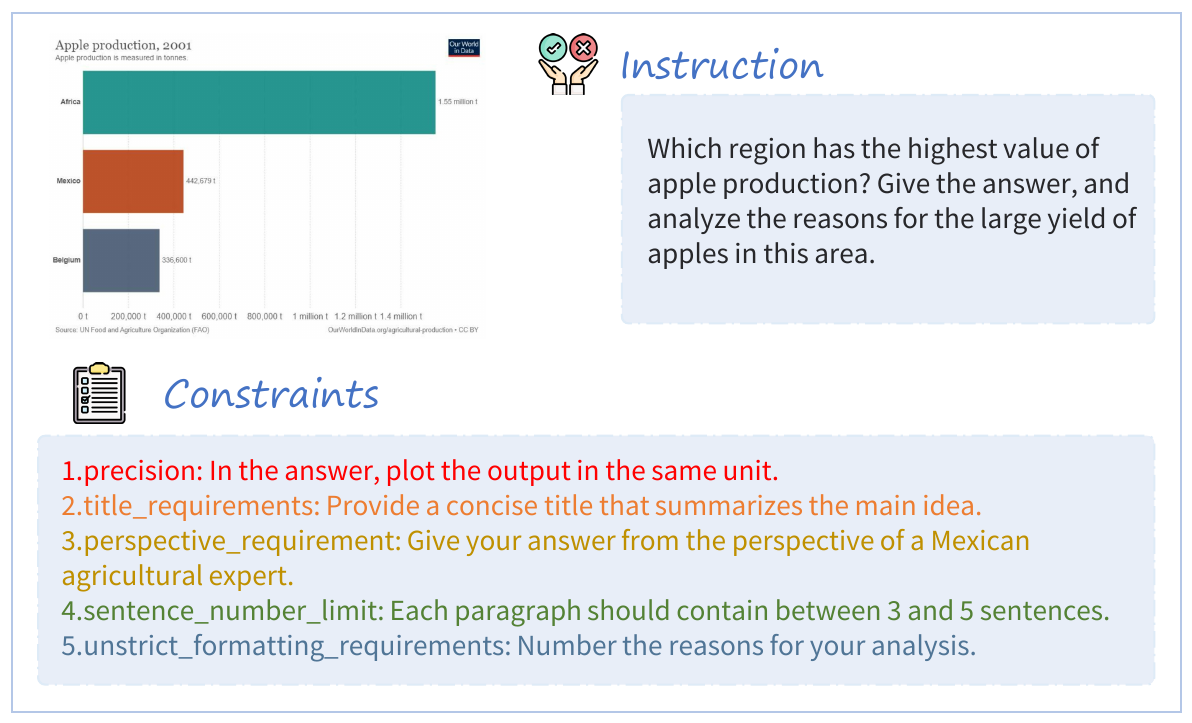}
    \caption{\textbf{A compose-level problem example from the MM-IFEval benchmark in the chart image category. }}
    \label{fig:t_chart}
    \vspace{-10pt}
\end{figure*}
\begin{figure*}[htbp]
    \centering
    \includegraphics[width=\linewidth]{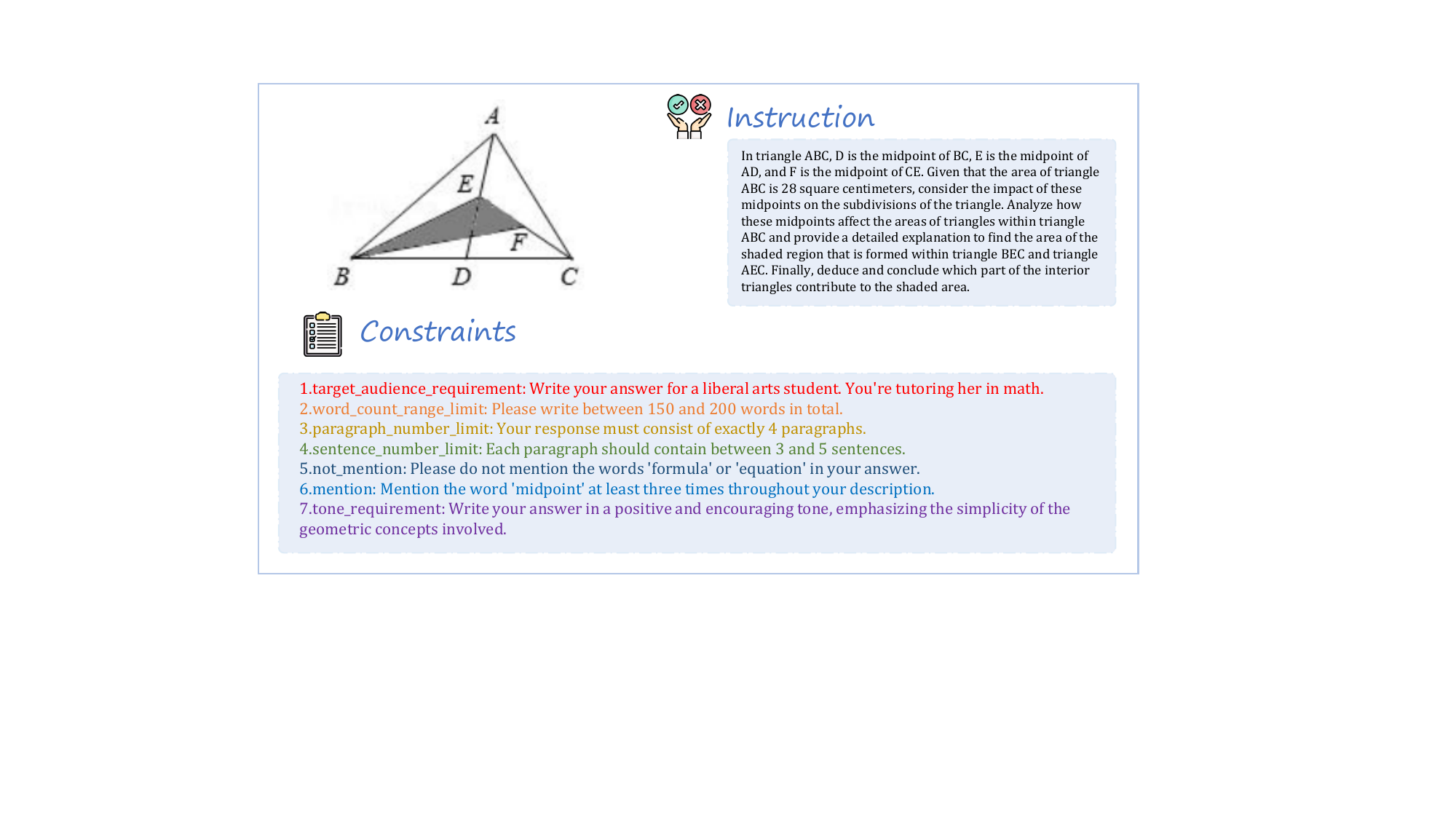}
    \caption{\textbf{A compose-level problem example from the MM-IFEval benchmark in the geometry image category. }}
    \label{fig:t_geo}
    \vspace{-10pt}
\end{figure*}


\clearpage 
\begin{table*}[p] 
\centering 
\setlength{\tabcolsep}{3pt} 
\tiny 
\begin{tabular}{|>{\raggedright\arraybackslash}p{1.6cm}|p{2.2cm}|p{1.3cm}|p{3.3cm}|p{3.3cm}|}
\hline
\textbf{Main Class} & \textbf{Subclass} & \textbf{Evaluation} & \textbf{Description} & \textbf{Example} \\
\hline
\multirow{2}{*}{\textbf{A. Rhetoric \& Logic}} & A.1 Rhetoric requirements &  Compare Judge & Constraint that requires the response to use a specific rhetorical technique. & ``Your output should include a metaphor." \\
\cline{2-5}
 & A.2 Logical relation &  Direct Judge & Constraint that ensures logical cohesion within the response by requiring specific logical connectors or structures. & ``Each paragraph must contain at least one cause-and-effect relationship." \\
\hline
\multirow{14}{*}{\textbf{B. Format limit}} & B.1 Natural language &  Direct Judge & Constraint specifying which natural language(s) should be used in the response. & ``Please answer in Spanish." \\
\cline{2-5}
 & B.2 Part of speech &  Direct Judge & Constraint that requires the response to use a specific part of speech. & ``Use at least three adjectives in your response." \\
\cline{2-5}
 & B.3 Sentence structure &  Direct Judge & Constraint that specifies special sentence structures to be used in the response. & ``Write each sentence so it includes a parenthetical phrase." \\
\cline{2-5}
 & B.4 Tense requirements &  Direct Judge & Constraint that specifies the use of multiple tenses within the response. & ``In past tense totally." \\
\cline{2-5}
 & B.5 Punctuation & Rule-base & Constraint specifying unconventional yet feasible punctuation usage in the response. & ``Replace all periods with semicolons." \\
\cline{2-5}
 & B.6 Highlight &  Direct Judge & Constraint that specifies a unique but manageable method for highlighting text. & ``Use **bold** for every noun." \\
\cline{2-5}
 & B.7 Title requirements &  Direct Judge & Constraint that specifies how titles should be added to the response. & ``Provide a concise title that summarizes the main idea." \\
\cline{2-5}
 & B.8 Style requirements &  Compare Judge & Constraint that specifies an unconventional or distinctive writing style for the response. & ``Write the answer in the form of a brief detective story." \\
\cline{2-5}
 & B.9 Case requirements &  Direct Judge & Constraint specifying an unusual yet readable approach to letter case in the response. & ``Write all nouns in UPPERCASE and all adjectives in lowercase." \\
\cline{2-5}
 & B.10 Unstrict format &  Direct Judge & Constraint specifying a unique format for the output while keeping it approachable. & ``Format your response as a short play script with speaker labels." \\
\cline{2-5}
 & B.11 Strict format &  Direct Judge & Constraint that requires the response to follow a strictly defined format. & ``Please provide the output as well-formed XML with custom tags." \\
\cline{2-5}
 & B.12 Number and List &  Direct Judge & Constraint for using numbered or bulleted lists in the response. & ``Present all key points as a numbered list with bulleted sub-lists." \\
\cline{2-5}
 & B.13 Wrap up &  Direct Judge & Constraint that requires a concise, well-structured summary or conclusion. & ``Provide a final paragraph summarizing the key arguments." \\
\cline{2-5}
 & B.14 First letter &  Direct Judge & Constraint specifying a pattern for the first letters of sentences or paragraphs. & ``Each sentence should begin with a letter that progresses through the alphabet." \\
\hline
\multirow{3}{*}{\textbf{C. Text Length limit}} & C.1 Paragraph limit & Rule-base & Constraint that specifies the number of paragraphs in the response. & ``Your response must consist of exactly 4 paragraphs." \\
\cline{2-5}
 & C.2 Sentence limit & Rule-base & Constraint that specifies the number of sentences in each paragraph. & ``Totally use 5 sentences in your response." \\
\cline{2-5}
 & C.3 Word limit & Rule-base & Constraint that specifies a small range for the total number of words in the text. & ``Your response must be a single word or phrase." \\
\hline
\multirow{2}{*}{\textbf{D. Math limit}} & D.1 Precision & Rule-base & Constraint that specifies the level of precision required in mathematical calculations. & ``Keep two decimal places for all numbers in the answer." \\
\cline{2-5}
 & D.2 Scientific notation & Rule-base & Constraint that requires the use of scientific notation for large or small numbers. & ``Express all numbers greater than 1,000 in scientific notation." \\
\hline
\multirow{7}{*}{\textbf{E. Action limit}} & E.1 Role imitation &  Compare Judge & Constraint requiring the response to imitate the tone and style of a specific role or public figure. & ``Please answer in the style of a sports commentator." \\
\cline{2-5}
 & E.2 Prefix and Suffix & Rule-base & Constraint that requires the response to begin or end with a specific phrase or symbol. & ``Please start your answer with 'Once upon a time...'." \\
\cline{2-5}
 & E.3 Tone requirement &  Compare Judge & Constraint specifying an emotional tone for the response. & ``Write your answer in a positive and encouraging tone." \\
\cline{2-5}
 & E.4 Perspective &  Direct Judge & Constraint that specifies a narrative perspective for the response. & ``Write your answer in the first-person singular as a personal account." \\
\cline{2-5}
 & E.5 Target audience &  Compare Judge & Constraint requiring the response to be tailored for a specific audience. & ``Craft your response as if explaining to high school students." \\
\cline{2-5}
 & E.6 Situation &  Compare Judge & Constraint requiring the response to be set in a specific situation or scenario. & ``Answer as if you are giving safety instructions before a flight." \\
\cline{2-5}
 & E.7 Prior condition &  Direct Judge & Constraint stating that when a specific condition is met, the response must follow a particular process. & ``If the user requests legal advice, begin with a disclaimer." \\
\hline
\multirow{4}{*}{\textbf{F. Keyword}} & F.1 Mention & Rule-base \&  Direct Judge & Constraint that requires including a specific keyword a certain number of times. & ``Mention 'GreenTech' exactly three times throughout." \\
\cline{2-5}
 & F.2 Not mention & Rule-base \&  Direct Judge & Constraint that requires avoiding specific keywords or phrases. & ``Do not mention the words 'budget' or 'investment'." \\
\cline{2-5}
 & F.3 Multiple mention & Rule-base \&  Direct Judge & Constraint requiring including multiple specified keywords in a balanced manner. & ``Mention both 'sustainability' and 'renewable energy' at least twice." \\
\cline{2-5}
 & F.4 Keyword variation &  Direct Judge & Constraint requiring the use of synonyms or variations of a given keyword. & ``Use at least three synonyms for 'innovation' throughout your text." \\
\hline
\end{tabular}
\caption{Constraint Categories and Evaluation Methods for MM-IFEval} 
\label{tab:subconstraints} 
\end{table*}
\clearpage 
\begin{table*}[t]
\label{tab:instruction_pool}
\centering
\resizebox{.7\textwidth}{!}{
\begin{tabular}{|m{3.5cm}|m{11cm}|}
\hline
Category & Instruction \\
\hline
\multirow{4}{*}{\centering Descriptive Analysis} & Describe the animal's typical habitat, diet, and one unique behavioral trait. \\
\cline{2-2}
 & Provide a detailed analysis of the image, including the setting, characters, and notable objects. \\
\cline{2-2}
 & Explain the activity taking place in the image. \\
\cline{2-2}
 & Describe the activities of the person on the left in the image. \\
\hline
\multirow{3}{*}{\centering Emotional \& Perspective} & What emotions do you think the person in this image might be feeling? \\
\cline{2-2}
 & Imagine you are the person on the left in the scene depicted in this image, write a story about what you would do next. \\
\cline{2-2}
 & Personify the sign in the image and express its feelings about the rule it presents. \\
\hline
\multirow{5}{*}{\centering Creative Writing} & Create a short conversation between any two individuals in the scene. \\
\cline{2-2}
 & Pretend this snapshot belongs to a larger story. Write a quick paragraph setting up the next plot twist. \\
\cline{2-2}
 & Use this picture as your muse. Craft a brief poem—any style—that captures the emotion you sense. \\
\cline{2-2}
 & Turn this scene into a short children's story focusing on wonder and curiosity. \\
\cline{2-2}
 & Write a short poem with two stanzas, inspired by the emotion or content depicted in this image. \\
\hline
\multirow{3}{*}{\centering Social Media \& Content} & Assume this is an image you are about to post on Twitter. Please provide a short, upbeat caption describing it. \\
\cline{2-2}
 & Assume you are creating a Pinterest pin with this image. Write a short inspirational or motivational caption to accompany it. \\
\cline{2-2}
 & If this image were promoting an upcoming event, compose a quick announcement with the date, a highlight of what to expect, and a call-to-action. \\
\hline
\centering Role Play & Imagine you are the photographer who took this picture. Briefly explain why you chose to capture this particular moment and what story you hope it conveys. \\
\hline
\end{tabular}
}
\caption{Task Pool for MM-IFEngine}
\label{tab:task_pool}
\end{table*}
\begin{table*}[htbp]
\centering
\scriptsize
\begin{tabular}{|>{\raggedright\arraybackslash}p{3.5cm}|>{\raggedright\arraybackslash}p{2.7cm}|p{4.2cm}|p{2.2cm}|}
\hline
\textbf{Verified Function Name} & \textbf{Function Parameters} & \textbf{Constraint Example} & \textbf{Parameter Example} \\
\hline
check\_whether\_ response\_paragraph\_ number\_in\_range & lower\_bound:int,\newline upper\_bound:int & The number of text paragraphs be at least 3 & [3, 10000] \\
\hline
check\_whether\_ response\_sentence\_ number\_in\_range & lower\_bound:int,\newline upper\_bound:int & The number of sentences be exactly 3 & [3, 3] \\
\hline
check\_whether\_each\_ paragraph\_sentence\_ number\_in\_range & lower\_bound:int,\newline upper\_bound:int & The number of sentences in each paragraph be less than 3 & [0, 2] \\
\hline
check\_whether\_each\_ paragraph\_sentence\_ number\_in\_range\_list & ranges:List[tuple] & The number of sentences in the first paragraph be exactly 3, and in the second paragraph be at most 2 & [[(3, 3), (1, 2)]] \\
\hline
check\_whether\_each\_ paragraph\_sentence\_ number\_exceeds & exceed\_num:int,\newline upper\_bound:int & Each new paragraph should have 1 sentence more than the previous one, no paragraph exceeds 7 sentences & [1, 7] \\
\hline
check\_whether\_ response\_word\_count\_ in\_range & lower\_bound:int,\newline upper\_bound:int & The number of words should be between 50 and 80 & [50, 80] \\
\hline
check\_whether\_each\_ paragraph\_word\_count\_ in\_range & lower\_bound:int,\newline upper\_bound:int & The number of words in each paragraph should be between 50 and 80 & [50, 80] \\
\hline
check\_whether\_each\_ paragraph\_word\_count\_ in\_range\_list & ranges:List[tuple] & The number of words in the first paragraph be between 20 and 30, in the second between 50 and 80 & [[(20, 30), (50, 80)]] \\
\hline
check\_whether\_whole\_ response\_not\_contain\_ certain\_substring & substring:str & The response should not contain the word "apple" & ["apple"] \\
\hline
check\_whether\_whole\_ response\_not\_contain\_ certain\_substrings & substrings:List[str] & The response should not contain the words "apple" and "banana" & [["apple", "banana"]] \\
\hline
check\_whether\_each\_ sentence\_begin\_with\_ certain\_substring & substring:str & Each sentence should start with exclamation point & ["!"] \\
\hline
check\_whether\_each\_ sentence\_end\_with\_ certain\_substring & substring:str & Each sentence should end with "apple" & ["apple"] \\
\hline
check\_whether\_whole\_ response\_begin\_with\_ certain\_substring & substring:str & The response should start with "apple" & ["apple"] \\
\hline
check\_whether\_whole\_ response\_end\_with\_ certain\_substring & substring:str & The response should end with "apple" & ["apple"] \\
\hline
check\_whether\_keywords\_ metioned\_in\_range & keywords:List[str],\newline lower\_bound\_times:int,\newline upper\_bound\_times:int & The response should mention the word "apple" at least 3 times & [["apple"], 3, 10000] \\
\hline
check\_number\_precision\_ in\_response & precision:int & The numbers in the response should have 2 decimal places & [2] \\
\hline
check\_whether\_has\_no\_ number\_in\_response & - & The response should not contain any number & [] \\
\hline
check\_scientific\_notation\_ precision\_in\_response & significant\_digits:int & The numbers in the response should have 3 significant digits & [3] \\
\hline
\end{tabular}
\caption{Verification Functions for rule-based evaluation method in MM-IFEval}
\label{tab:verification_functions}
\end{table*}

\end{document}